\documentclass[journal]{IEEEtran}
\usepackage[pdftex]{graphicx}
\usepackage{amsfonts}
\usepackage{tabularx}
\usepackage{multirow}
\usepackage[ruled,vlined]{algorithm2e}
\usepackage{makecell}
\usepackage{{tabularx}}
\usepackage{epstopdf}
\usepackage{booktabs}
\usepackage{threeparttable}
\usepackage{mathrsfs}
\usepackage{threeparttable}
\usepackage{color}

\ifCLASSINFOpdf
\else
\fi
\begin{document}
%
\title{Cross-Database Micro-Expression Recognition: A Benchmark}
%
%
%

\author{Yuan~Zong,~\IEEEmembership{Member,~IEEE,}
Tong~Zhang$^*$,~\IEEEmembership{Member,~IEEE,}
Wenming~Zheng$^*$,~\IEEEmembership{Senior~Member,~IEEE,}
Xiaopeng~Hong,
Chuangao~Tang,
Zhen~Cui,~\IEEEmembership{Member,~IEEE,}
and~Guoying~Zhao,~\IEEEmembership{Senior~Member,~IEEE}
\thanks{Y. Zong, W. Zheng, and C. Tang are with the Key Laboratory of Child Development and Learning Science of Ministry of Education, School of Biological Sciences and Medical Engineering, Southeast University, Nanjing 210096, China. Email: \{xhzongyuan, wenming\_zheng, tcg2016\}@seu.edu.cn}
\thanks{T. Zhang is with the School of Computer Science and Engineering, South China University of Technology, Guangzhou 510006, China. Email: tony@scut.edu.cn}
\thanks{X. Hong is with the Xi'an Jiao Tong University, Xi'an 710049, China. Email: hongxiaopeng@ieee.org}
\thanks{G. Zhao is with the Center for Machine Vision and Signal Analysis, University of Oulu, Finland. Email: \{xiaopeng.hong, guoying.zhao\}@oulu.fi}
\thanks{Z. Cui is with the School of Computer Science and Engineering, Nanjing University of Science and Technology, Nanjing 210094, China. Email: zhen.cui@njust.edu.cn}
\thanks{The asterisks indicate the corresponding authors.}
}

\maketitle

\begin{abstract}
Cross-database micro-expression recognition (CDMER) is one of recently emerging and interesting problem in micro-expression analysis. CDMER is more challenging than the conventional micro-expression recognition (MER), because the training and testing samples in CDMER come from different micro-expression databases, resulting in the inconsistency of the feature distributions between the training and testing sets. In this paper, we contribute to this topic from three aspects. First, we establish a CDMER experimental evaluation protocol aiming to allow the researchers to conveniently work on this topic and provide a standard platform for evaluating their proposed methods. Second, we conduct benchmark experiments by using NINE state-of-the-art domain adaptation (DA) methods and SIX popular spatiotemporal descriptors for respectively investigating CDMER problem from two different perspectives. Third, we propose a novel DA method called region selective transfer regression (RSTR) to deal with the CDMER task. Our RSTR takes advantage of one important cue for recognizing micro-expressions, i.e., the different contributions of the facial local regions in MER. The overall superior performance of RSTR demonstrates that taking into consideration the important cues benefiting MER, e.g., the facial local region information, contributes to develop effective DA methods for dealing with CDMER problem.
\end{abstract}


%
\IEEEpeerreviewmaketitle

\section{Introduction}
%
%
%
%
\IEEEPARstart{M}{icro-expression} is one involuntary facial expression whose duration is usually within $\frac1{25}$ seconds \cite{ekman1969nonverbal}. In 1966, Haggard and Isaacs~\cite{haggard1966micromomentary} first discovered micro-expressions during their research of ego mechanisms in psychotherapy. Subsequently, Ekman et al.~\cite{ekman1969nonverbal} found this type of facial expressions again when they observed a video of a psychotic patient and formally named them micro-expressions. Different from ordinary facial expressions, micro-expressions happen as a result of conscious suppression or unconscious repression. They can be viewed as a ``leakage'' often occurring on someone's face when that person tries to conceal a genuine emotion. In other words, people without highly professional training cannot hide micro-expressions and thus micro-expression can usually expose people's true emotions~\cite{ekman2009telling}. For this reason, understanding micro-expressions has great values in lots of practical applications, e.g., lie detection. An interrogator who is good at recognizing micro-expressions is able to spot the discrepancies between what he/she hear and what he/she see from the criminal suspect.

Unfortunately, without proper training most people cannot recognize micro-expressions in real time. In order to lower this barrier, researchers from computer vision and affective computing community have been focusing on automatic micro-expression recognition (MER) techniques and proposed various methods~\cite{pfister2011recognising,wang2014lbp,wang2015micro,liu2016main,kim2016micro,lu2016micro,xu2017microexpression,happy2017fuzzy,zong2018learning}. For example, Pfister et al.~\cite{pfister2011recognising} adopted the local binary pattern from three orthogonal planes (LBP-TOP)~\cite{zhao2007dynamic} to describe micro-expressions and demonstrated the effectiveness of spatiotemporal descriptors in MER. Since then, lots of excellent spatiotemporal descriptors have been designed for describing micro-expressions, e.g., LBP with six intersection points (LBP-SIP)~\cite{wang2014lbp}, facial dynamics map (FDM)~\cite{xu2017microexpression}, and fuzzy histogram of optical flow orientation (FHOFO)~\cite{happy2017fuzzy}. Besides the study of spatiotemporal descriptors, some important cues of micro-expression samples have been explored for developing effective MER methods. In the work of~\cite{wang2015micro}, Wang et al. investigated whether the color information is beneficial for MER by proposing a tensor independent color space (TICS) method to decompose micro-expression samples into different color channels. Inspired by the work of facial action coding system (FACS)\footnote{According to the definition of Ekman et al.~\cite{ekman1997face}, FACS contains lots of AUs and action descriptors (ADs). With AUs and ADs, we are able to code human beings' basic emotions, e.g., happy and disgust.}~\cite{ekman1997face}, Wang et al.~\cite{wang2015micro} and Liu et al.~\cite{liu2016main} respectively designed a set of Region-of-Interests (ROIs) which covers one or more micro-expression aware action unit (AU) regions and leveraged these AU region information to deal with MER problem. Recently, deep learning techniques are also used for handling MER. For instance, Kim et al.~\cite{kim2016micro} made use of well-performing convolutional neural networks (CNNs)~\cite{krizhevsky2012imagenet} and long short-term memory (LSTM) recurrent neural networks~\cite{hochreiter1997long} to design a straightforward feature learning network for MER tasks.

Although the above MER methods have achieved promising performance, it should be pointed out that the above methods are still far from the requirements of a high-quality MER system. One major reason is that existing MER methods are mostly designed and evaluated without the consideration of the complex scenarios encountered by the MER system in practice. For example, the training and testing micro-expression samples provided for MER system may be recorded by different cameras (e.g., high-speed camera v.s. near-infrared camera) or under different environments (e.g., normal illumination v.s. weak illumination). In this scenario, the performance of the above MER methods may sharply drop due to the largely different feature distributions existing between the training and testing micro-expression samples caused by the heterogeneous video qualities. It thus brings us a new topic in micro-expression analysis, i.e., \textbf{cross-database micro-expression recognition (CDMER)}, in which the training and testing samples come from two different micro-expression databases collected by different cameras or under different environments. CDMER offers a good way to mimic the scenarios the MER system would encounter in reality. Therefore, it is worthy to deeply investigate this challenging topic.

Similar with MER, the cross-database recognition problems have been extensively studied in many other modalities for emotion recognition such as speech emotion recognition (SER)~\cite{schuller2010cross}, facial expression recognition (FER)~\cite{yan2016cross}, and EEG emotion recognition~\cite{zheng2016personalizing}. However, there are several limitations existing in current cross-database emotion recognition research. First, in cross-database emotion recognition, there is a lack of unified standard evaluation protocol. Researchers often choose their preferred experiment materials including emotion databases, emotion features, classifiers, and evaluation metrics to set up their own experimental evaluation protocol. It raises the barriers of entry to this topic because it would cost lots of time to prepare evaluation protocol for the new researchers who are interested in but not familiar with this topic. Second, it can be found that cross-database emotion recognition was purely viewed as a domain adaptation (DA)~\cite{pan2010survey} task in almost all of the existing works. For example, in the works of~\cite{zong2017learning,zong2018domain}, Zong et al. investigated CDMER completely from the perspective of DA and designed a series of DA methods to deal with it. It should be noted that cross-database emotion recognition including CDMER is not  a simple DA problem and is strongly dependent on the micro-expression features. Designing an excellent and robust feature used for describing micro-expressions would improve the recognition performance of classifier in the CDMER tasks as well, which provides another feasible solution for CDMER problem.

Based on the above considerations, in this paper we investigate the CDMER problem to break through the above two limitations widely existing in cross-database emotion recognition research. To this end, we made three contributions in this paper as follows:
\begin{enumerate}
\item We build a CDMER experimental evaluation protocol and design a set of CDMER experiments based on two public available micro-expression databases, which can be served as a standard platform for evaluating the CDMER methods from not only the perspective of DA but also the micro-expression features.

\item We use NINE representative DA methods and SIX spatiotemporal descriptors used for describing micro-expressions to conduct extensive benchmark evaluation experiments under the designed protocol and deeply disucss the experimental results.

\item To deal with CDMER problem, we also propose a novel DA method called region selective transfer regression (RSTR), which takes the facial local region information of micro-expressions into full consideration.
\end{enumerate}

It should be pointed out that this work is the extended version of our previous conference paper~\cite{zong2019cross}. The major motivation of this work is to attract and encourage more researchers to join this challenging but interesting topic and provide convenience for them to get started. For this reason, we released all the data and codes involving CDMER in this paper on our project website: http://aip.seu.edu.cn/cdmer.

The rest of this paper is organized as follows: Section~\ref{benchmark} describes the details of the CDMER benchmark including the designed experimental evaluation protocol and the evaluated methods. Section~\ref{RSTR} introduces the proposed RSTR. The experimental results are reported and discussed in Section~\ref{experiment}. Finally, the paper is concluded in Section~\ref{conclusion}.

\begin{figure*}[!t]
\centering
\includegraphics[width=0.8\textwidth]{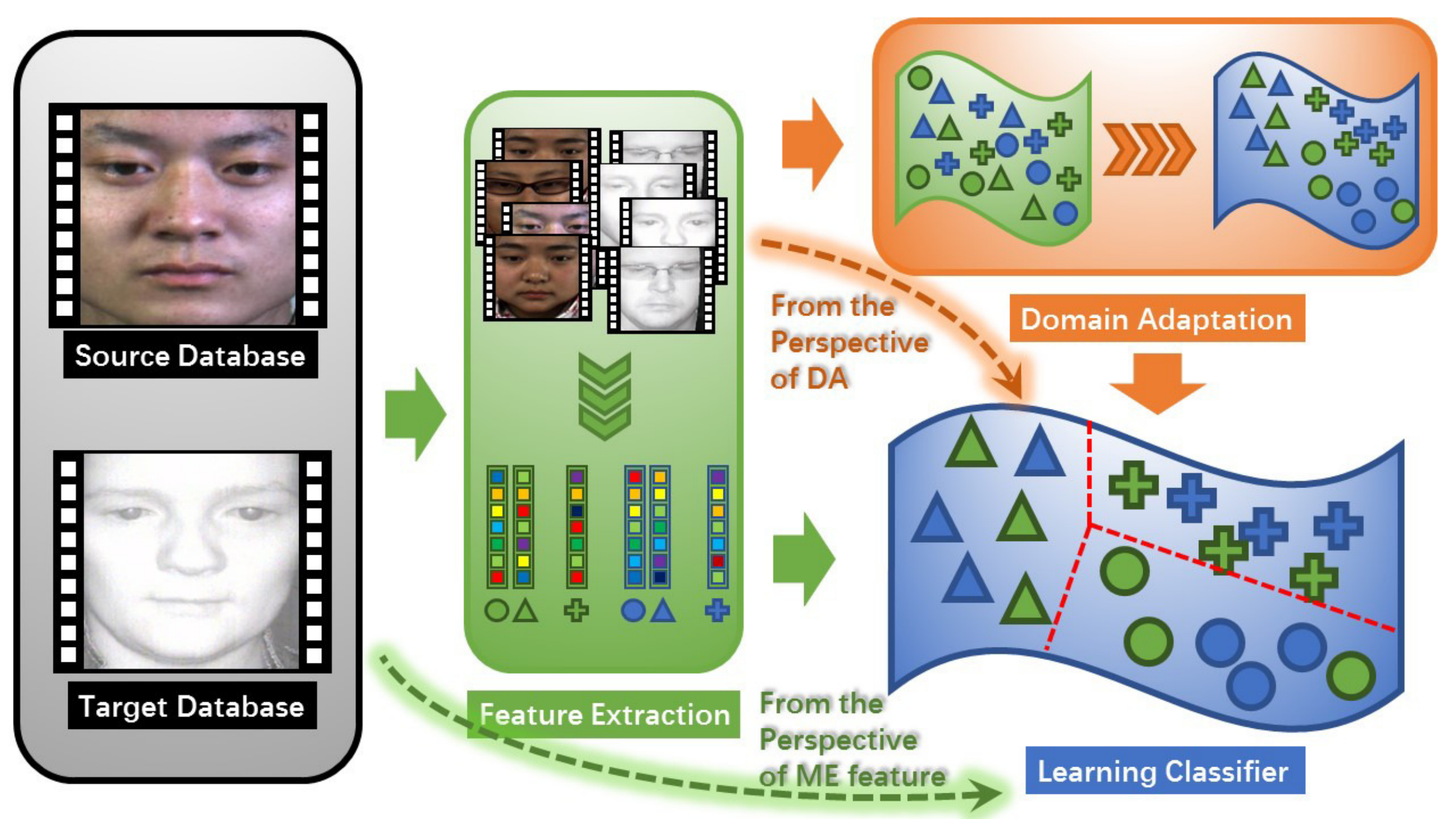}
\caption{An illustration of how to solve the CDMER problem from the perspectives of DA methods and micro-expression features, respectively. The DA solution for CDMER problem is demonstrated as the direction of orange dash line. What the green dash line directs shows the solution from the micro-expression feature perspective. Micro-expression feature solution targets at designing the robust features that are less sensitive to the database variance for describing micro-expressions. In this case, CDMER consists of two major steps, i.e., feature extraction and classifier learning. Apart from feature extraction and classifier learning, the DA solution has an additional step between the steps of feature extraction and classifier learning, i.e., leveraging the DA techniques to narrow the feature distribution gap between the source and target data in the original feature space.}
\label{fig:fig1}
\end{figure*}

\section{Benchmark Detail}
\label{benchmark}

As described previously, most existing cross-database emotion recognition problems including CDMER are often viewed as a DA task and solved by DA methods. In this way, followed by feature extraction, DA technique is used to relieve the feature distribution mismatch between the source and target micro-expression samples. Then, we are able to learn a classifier based on the labeled source micro-expression database to predict the micro-expression categories of samples from target database. A picture is drawn to illustrate the detailed process of using DA methods to deal with the CDMER problem, which is shown following the sequence of the orange dash arrow line in Fig.~\ref{fig:fig1}. It should be pointed out that besides DA solution, designing robust (database invariant) micro-expression features is also an effective way to solve CDMER problem. By resorting to well-designed robust micro-expression features, CDMER can be actually solved as a typical pattern recognition task which only needs two major steps including feature extraction and classifier learning. We draw its detail in what the green dash arrow directs in Fig.~\ref{fig:fig1}. Following the ideas of these two solutions for CDMER, we would like to design a CDMER evaluation protocol which can be served as a standard platform used for evaluating both DA methods and micro-expression features, respectively.

\subsection{A Standard Evaluation Protocol for CDMER}

\subsubsection{Data Preparation}

Two publicly available spontaneous micro-expression databases are adopted for building the benchmark evaluation experiments, i.e., CASME~II~\cite{yan2014casme} and SMIC~\cite{li2013spontaneous}. CASME~II was built by Yan et al. from Institute of Psychology, Chinese Academy of Sciences. It consists of 257 micro-expression samples from 26 subjects. Among these 257 samples, the micro-expression label of one sample is not provided. Each of the other 256 samples is assigned one of seven micro-expression labels including $Happy$, $Disgust$, $Repression$, $Surprise$, $Sad$, $Fear$, and $Others$. Different from CASME~II, Li et al. from University of Oulu, Finland considered the image quality diversity of micro-expression samples and hence employed three cameras, i.e., a high-speed (HS) camera, a normal visual (VIS) camera, and a near-infrared (NIR) camera, to collect three subsets to obtain the SMIC (HS, VIS, and NIR) database. The HS subset has 164 samples belonging to 16 subjects and 71 samples of eight subjects from these 16 subjects compose VIS and NIR subsets. All the samples in SMIC are categorized into three types of micro-expressions, i.e., $Positive$, $Negative$, and $Surprise$. To make CASME~II and SMIC have the same micro-expression labeling, we select the samples of $Happy$, $Disgust$, $Surprise$, $Sad$, and $Fear$ from CASME~II and relabel them according to the labeling rule in SMIC, where $Happy$ samples are given the $Positive$ labels, $Disgust$, $Sad$, and $Fear$ samples are relabeled with $Negative$ micro-expression, and the labels of $Surprise$ sample keep unchanged. The sample constitution with respect to consistent categories of the selected CASME~II and SMIC databases is shown in Table~\ref{tab:tab1}.

\begin{table}[!t]
\scriptsize
\renewcommand{\arraystretch}{1.2}
\caption{The Sample Constitutions of the selected CASME~II and SMIC Databases with the Same Micro-Expression Labels for CDMER.}
\label{tab:tab1}
\centering
\begin{tabular}{|l|c|c|c|}
\hline
Micro-Expression Database & Positive & Negative & Surprise \\ \hline\hline
Selected CASME~II & 32 & 73 & 25 \\ \hline
SMIC (HS) & 51 & 70 & 43 \\ \hline
SMIC (VIS) & 23 & 28 & 20 \\ \hline
SMIC (NIR) & 23 & 28 & 20 \\ \hline
\end{tabular}
\end{table}

\subsubsection{CDMER Tasks}

Following our preliminary works of~\cite{zong2017learning,zong2018domain}, we design two kinds of CDMER tasks based on the selected CASME~II and SMIC databases for our CDMER protocol. The first type of tasks denoted by TYPE-I is the one between either two datasets of SMIC (HS, VIS, and NIR). The second type of tasks chooses the selected CASME~II and one dataset of SMIC (HS, VIS, and NIR) to serve as source and target micro-expression databases, alternatively, which is denoted by TYPE-II. This leads to totally 12 CDMER experiments, where one CDMER task is denoted by $Exp.i: S \rightarrow T$, where $Exp.i$ is the number of this experiment and $S$ and $T$ are the source and target micro-expression databases, respectively. We summarize all the CDMER experiments in the designed protocol in Table~\ref{tab:tab2}.

\begin{table}[!t]
\scriptsize
\renewcommand{\arraystretch}{1.2}
\caption{The Detailed Information of Two Types of CDMER Tasks in the Designed Evaluation Protocol.}
\label{tab:tab2}
\centering
\begin{tabular}{|c|c|c|c|}
\hline
TYPE & CDMER Task & Source Database & Target Database\\ \hline\hline
\multirow{6}{*}{TYPI-I} & $Exp.1: H \rightarrow V$ & SMIC (HS) & SMIC (VIS) \\ \cline{2-4}
{} & $Exp.2: V \rightarrow H$ & SMIC (VIS) & SMIC (HS) \\ \cline{2-4}
{} & $Exp.3: H \rightarrow N$ & SMIC (HS) & SMIC (NIR) \\ \cline{2-4}
{} & $Exp.4: N \rightarrow H$ & SMIC (NIR) & SMIC (HS) \\ \cline{2-4}
{} & $Exp.5: V \rightarrow N$ & SMIC (VIS) & SMIC (NIR) \\ \cline{2-4}
{} & $Exp.6: N \rightarrow V$ & SMIC (NIR) & SMIC (VIS) \\ \hline \hline
\multirow{6}{*}{TYPE-II} & $Exp.7: C \rightarrow H$ & Selected CASME~II & SMIC (HS) \\ \cline{2-4}
{} & $Exp.8: H \rightarrow C$ & SMIC (HS) & Selected CASME~II \\ \cline{2-4}
{} & $Exp.9: C \rightarrow V$ & Selected CASME~II & SMIC (VIS) \\ \cline{2-4}
{} & $Exp.10: V \rightarrow C$ & SMIC (VIS) & Selected CASME~II \\ \cline{2-4}
{} & $Exp.11: C \rightarrow N$ & Selected CASME~II & SMIC (NIR) \\ \cline{2-4}
{} & $Exp.12: N \rightarrow C$ & SMIC (NIR) & Selected CASME~II \\ \hline
\end{tabular}
\end{table}

\subsubsection{Performance Metrics}

In our previous works of~\cite{zong2017learning,zong2018domain}, weighted average recall (WAR) and unweighted average recall (UAR) are employed to serve as the performance metrics, where WAR is the normal recognition $Accuracy$ while UAR is the mean accuracy of each class divided by the number of the classes without the consideration of sample number of each class. The main reason of introducing UAR is due to the class imbalanced problem which widely exists in CASME~II and SMIC databases. As shown in Table~\ref{tab:tab1}, the number of $Negative$ samples in the selected CASME~II is 73, which is significantly larger than the numbers of the remaining two types of micro-expression samples (32 for $Positive$ and 25 for $Surprise$).

$Mean$ $F1$-$score$ is another recommended metric, which has been widely used to avoid the bias in performance measurement caused by the class imbalanced problem in MER literatures~\cite{le2014spontaneous,oh2015monogenic,le2016sparsity,le2016eulerian,zong2018learning}. For this reason, in our benchmark, we adopt the combination of $mean$ $F1$-$score$ and $Accuracy$ to serve as the metrics, where $mean$ $F1$-$score$ is the main metric and recognition accuracy  as the secondary one. The $mean$ $F1$-$score$ is calculated according to $mean$ $F1$-$score = \frac1c\sum_{i=1}^c\frac{2 p_i \times r_i}{p_i + r_i}$, where $p_i$ and $r_i$ mean the precision and recall of the $i^{th}$ micro-expression, respectively, and $c$ is the number of micro-expressions. The $Accuracy$ is calculated by $Accuracy = \frac{T}{N} \times 100$, where $T$ and $N$ are the number of correct predictions and the number of target micro-expression samples.

\subsubsection{Preprocessing and Feature Extraction}
\label{multi-scale}
\begin{figure}[!t]
\centering
\includegraphics[width=\columnwidth]{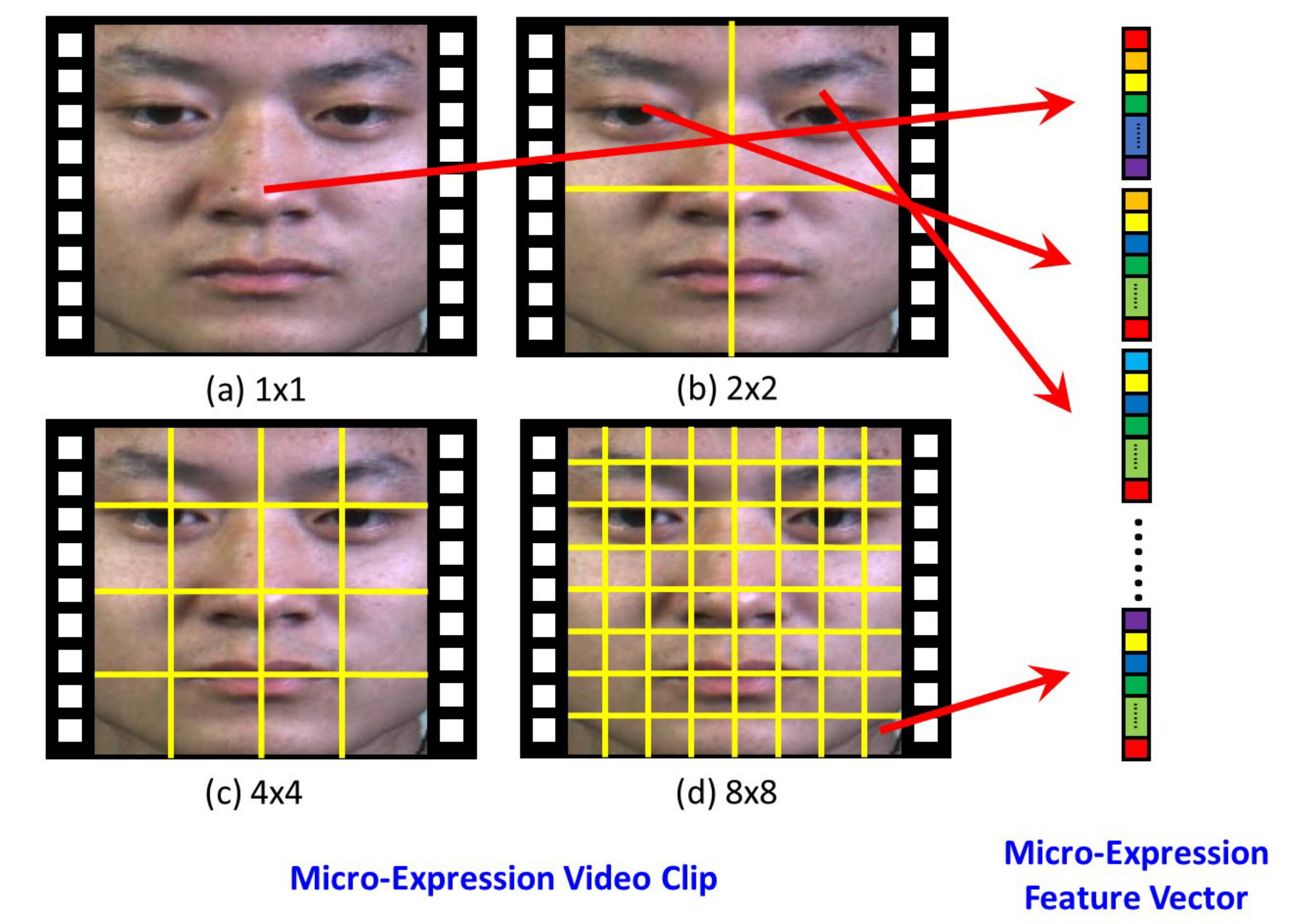}
\caption{Multi-Scale Grid Based Spatial Division Scheme for Micro-Expression Feature Extraction Used in the Benchmark.}
\label{fig:fig2}
\end{figure}

Before feature extraction, preprocessing operations, e.g., face alignment and face cropping, are performed on the micro-expression samples. For convenience, we directly adopted the image sequence data preprocessed by the collectors of CASME~II and SMIC for the benchmark evaluation experiments. Then, we employ the temporal interpolation model (TIM)~\cite{zhou2011towards} to normalize the frame number of all the micro-expression video clips to 16 and resize each frame image to $112\times112$, which allows us to extract spatiotemporal descriptor with specific parameter settings for micro-expression samples\footnote{For example, LBP-TOP with $R=3$ requires at least SEVEN frames for a given micro-expression video clip. However, some micro-expression video clip has SIX or less frames, e.g., in VIS subset of SMIC. Therefore, frame number normalization is a necessary preprocessing step when we choose some spatiotemporal descriptors to describe micro-expression.}. Furthermore, we compute the multi-scale spatiotemporal descriptors using four types of spatial grids ($1\times1, 2\times2, 3\times3,$ and $4\times4$) shown in Fig.~\ref{fig:fig2} to serve as the micro-expression features. This is expected to extensively cover micro-expression related facial local regions and increase the discriminative power of the extracted spatiotemporal descriptor~\cite{zong2018learning}. As Fig.~\ref{fig:fig2} shows, given a micro-expression sample $\mathcal{M}$, the spatiotemporal descriptor corresponding to each facial block denoted by $\mathbf{x}_i~(i=1,\cdots,K)$, where the facial block number $K = 85$, is first extracted one by one and then compose the final micro-expression feature vectors, which can be formulated as $\mathbf{x} = [\mathbf{x}_1,\cdots,\mathbf{x}_K]^T$.

As described previously, we attempt to investigate CDMER problem from two different perspectives including domain adaptation (DA) and micro-expression feature extraction. By using effective DA methods and micro-expression features, the large feature distribution difference between the source and target micro-expression databases in CDMER would be relieved. Hence, in our benchmark, we will respectively evaluate the performance of state-of-the-art DA methods and spatiotemporal descriptors used for describing micro-expressions in all the above 12 designed CDMER experiments. Note that in the experiments of DA methods, we suggest to employ a baseline spatiotemporal descriptor, uniform LBP-TOP~\cite{zhao2007dynamic} with fixed parameters (neighboring radius $R$ and number of the neighboring points $P$ for LBP operator on three orthogonal planes are fixed at 3 and 8, respectively), as shown in Fig.~\ref{fig:fig2} to serve as the micro-expression feature vector $\mathbf{x}^v$. In the case of using various features, we extract different $\mathbf{x}^v$ corresponding to different types of spatiotemporal descriptors used for describing micro-expression and then conduct CDMER experiments.

\subsubsection{Classifier}

To offer a fair comparison, the linear SVM is suggested for serving as the classifier in our benchmark. Without specific description, LibSVM~\cite{chang2011libsvm} is used in the implementation of SVM for both micro-expression features and DA evaluation experiments.

\subsection{Evaluated Methods}

\subsubsection{DA Methods}
For the evaluation of methods for dealing with CDMER from the perspective of DA, the following baseline and state-of-the-art unsupervised DA methods are employed.

\textbf{Baseline (SVM without DA)}~\cite{chang2011libsvm}: A support vector machine (SVM) without any domain adaptation is served as the baseline method. In the evaluation experiments, we directly learn the linear SVM on the source micro-expression database and then use it to predict the micro-expression labels of samples from the target database.

\textbf{IW-SVM}~\cite{hassan2013acoustic}: Importance-weighted SVM (IW-SVM) was proposed by Hassan et al. to deal with cross-database speech emotion recognition tasks. In this method, a transfer learning method, e.g., kernel mean matching (KMM)~\cite{huang2006correcting}, unconstrained least-squares importance fitting (uLSIF)~\cite{kanamori2009least}, and Kullback-Leibler importance estimation procedure~\cite{sugiyama2008direct}, is first used to learn a group of importance weights for source samples and then these weights are incorporated into the SVM classifier to eliminate the feature distribution difference between the source and target samples.

\textbf{TCA}~\cite{pan2011domain}: Transfer component analysis (TCA) was proposed by Pan et al., which is to seek some transfer components across domains in a reproducing kernel Hilbert space (RKHS). By using these transfer components, a subspace can be spanned in which the sample distributions from different domains would be close to each other.

\textbf{GFK}~\cite{gong2012geodesic}: Geodesic flow kernel (GFK) was proposed by Gong et al. and has been a widely-used baseline comparison method in DA research. GFK aims to bridge two domains and narrow their gaps with a well-designed geodesic flow kernel on a Grassmann manifold.

\textbf{SA}~\cite{fernando2013unsupervised}: Subspace alignment (SA) is another popular unsupervised DA method and have been served as the baseline comparison method in lots of DA literatures. The target of SA method is to seek a mapping function which can aligns the subspace the source samples lie in with respect to the target ones.

\textbf{STM}~\cite{chu2013selective,chu2017selective}: Selective transfer machine (STM) was originally proposed to cope with personalized facial action unit (AU) detection problem. STM makes use of an instance-wise weighted SVM to model the relationship between the training samples and its AU information and meanwhile KMM to eliminate the difference between the AU samples from testing subject and training subjects.

\textbf{TKL}~\cite{long2015domain}: Long et al. proposed a novel unsupervised DA method called transfer kernel learning (TKL), which aims to learn a domain invariant kernel for eliminating the feature distribution mismatch between the source and target domains.

\textbf{TSRG}~\cite{zong2017learning}: TSRG is short for target sample re-generator and was proposed to deal with CDMER problem. The aim of TSRG is to learn a sample regenerator for the target micro-expression samples and the feature distribution gap between the source and target micro-expression databases would be narrowed after the regeneration operation.

\textbf{DRFS-T}~\cite{zong2018domain}: TSRG is further extended to a generalized framework called domain regeneration (DR), which inherits the basic idea of TSRG. DRFS-T is one new designed sample regenerator under the DR framework and means domain regeneration in the original feature space with unchanged target samples and it shares the similar idea with
TSRG. Their only difference is that DRFS-T keeps the target samples unchanged and regenerates the source samples to have the same or similar feature distributions of target samples, while TSRG is opposite.

\textbf{DRLS}~\cite{zong2018domain}: DRLS is another newly designed sample regenerator based on the DR framework. Different from TSRG and DRFS-T, the subspace used in performing regeneration in DRLS is the label space spanned by the label information provided in the source micro-expression database instead of original feature space.

\subsubsection{Micro-Expression Features}

For exploring the performance of different existing micro-expression features in coping with CDMER problem, we collect following SIX representative handcrafted spatiotemporal descriptors and ONE deep spatiotemporal descriptor to conduct the benchmark evaluation experiments:

\textbf{LBP-TOP}~\cite{zhao2007dynamic}: LBP-TOP is an spatiotemporal extension of LBP by performing LBP coding on three orthogonal planes. It is originally proposed to deal with dynamic texture recognition tasks and recently has been widely used for describing micro-expressions~\cite{pfister2011recognising,wang2015micro,liu2016main,zong2018learning}.

\textbf{LBP-SIP}~\cite{wang2014lbp}: LBP-SIP is short for LBP with six intersection points. Wang et al. designed it in order to reduce the redundancy in LBP-TOP patterns and provide a more compact and lightweight representation. Compared with LBP-TOP, LBP-SIP uses six intersection points in the intersection lines surrounding the center points for LBP coding and hence its computational complexity is significantly reduced.

\textbf{LPQ-TOP}~\cite{paivarinta2011volume}: Following the manner of LBP-TOP, local phase quantization (LPQ)~\cite{ojansivu2008blur}, which quantifies the Fourier transform phase in local neighborhoods, is also extended to the spatiotemporal version called LPQ from three orthogonal planes (LPQ-TOP).

\textbf{HOG-TOP}~\cite{li2017towards}: Histograms of oriented gradients (HOG)~\cite{dalal2005histograms} is earliest proposed for human detection and subsequently applied on lots of vision tasks. In the work of~\cite{li2017towards}, Li et al. extends HOG to a 3D version called HOG-TOP, which borrows the basic idea of LBP-TOP.

\textbf{HIGO-TOP}~\cite{li2017towards}: Histogram of image gradient orientation (HIGO) was proposed in the work of~Li et al.~\cite{li2017towards} by degenerating HOG. Compared with HOG, HIGO simply uses vote rather than weighted vote in counting the responses of the histogram bins. As the name suggests, HIGO-TOP is the 3D extension of HIGO by using the manner of LBP-TOP.

\textbf{C3D}~\cite{tran2015learning}: Recently, the research of spatiotemporal deep feature learning models have made great progress. Three-dimensional convolutional neural network (C3D) is one of excellent representatives and has gained promising performance in video based action recognition tasks. C3D can be actually viewed as a 3D extension of VGG network~\cite{simonyan2014very}, which replaces 2D convolution and pooling operations with 3D ones.

Besides the above state-of-the-art DA methods and spatiotemporal descriptors, in this paper we also propose a novel DA method called \textbf{region selective transfer regression (RSTR)} for dealing with CDMER problem and evaluate its performance. The major advantage of the proposed RSTR is that we further consider the different contributions of the facial local regions in the design of RSTR. The detailed information of RSTR including the formulation and optimization is given in Section~\ref{RSTR}.

%
%
%

\section{Region Selective Transfer Regression}
\label{RSTR}
\subsection{Formulation}
Let $\mathbf{X}^s = [\mathbf{X}_1^{{s}^T},\cdots,\mathbf{X}_K^{{s}^T}]^T \in \mathbb{R}^{Kd\times N_s}$ and $\mathbf{X}^t = [\mathbf{X}_1^{{t}^T},\cdots,\mathbf{X}_K^{{t}^T}]^T \in \mathbb{R}^{Kd\times N_t}$ be the micro-expression feature matrices of the samples belonging to source and target micro-expression databases, where each column in $\mathbf{X}^s$ and $\mathbf{X}^t$ is the feature vector like $\mathbf{x}$ shown in Section~\ref{multi-scale}, $K$ is the number of the divided facial blocks, $N_s$ and $N_t$ are the source and target sample numbers, and $d$ is the dimension of the spatiotemporal descriptor vector extracted from each facial blocks, respectively. According to the problem setting of CDMER, it is known that the label information of the source micro-expression database is provided while the target micro-expression samples are entirely unlabeled. We hence assign source micro-expression database with a label matrix $\mathbf{L}^s$ whose $i^{th}$ column $\mathbf{l}_i^s$ reveals the micro-expression category of its corresponding source micro-expression sample in $\mathbf{X}^s$. $\mathbf{l}_i^s$ is a binary vector and the $c^{th}$ element is 1 if and only if it belongs to the $c^{th}$ micro-expression category.

Intuitively, we are able to use a simple linear regression model to build the relationship between the source features and their corresponding label information, which is formulated as follows:
\begin{eqnarray}
\min_\mathbf{\mathbf{C}} \Vert \mathbf{L}^s - \mathbf{C}^T\mathbf{X}^s \Vert_F^2,
\end{eqnarray}
where $\mathbf{C}$ is the regression coefficient matrix. It is clear that the learned regression coefficient matrix of the above problem can only suits the micro-expression recognition problem based on source database and cannot be applicable to the target micro-expression samples.

To overcome this shortcoming of the conventional regression model, we would like to extend it to a transfer regression model such that the learned regression coefficient matrix can also be suitable for the target micro-expression samples. To this end, we introduce the maximum mean discrepancy (MMD)~\cite{borgwardt2006integrating}, which measures the feature distribution distance between two data sets, for building our transfer regression model. MMD is defined as the distance between the centers of two different data sets in a kernel space and can be formulated as
\begin{eqnarray}
\mathbb{MMD}(\mathbf{X}^s,\mathbf{X}^t) = \Vert \frac1{N_s}\Phi(\mathbf{X}^s)\mathbf{1}_s - \frac1{N_s}\Phi(\mathbf{X}^t)\mathbf{1}_t \Vert_\mathcal{H},
\end{eqnarray}
where $\Phi$ is a kernel mapping operator. It had been proved that $\Phi(\mathbf{X}^s)$ and $\Phi(\mathbf{X}^t)$ would have the same or similar feature distributions if their MMD is close to 0 after arising dimension using $\Phi$. By using MMD as the regularization term, we are able to arrive at our transfer regression model which has the following formulation:
\begin{eqnarray}
\min_\mathbf{\mathbf{C},\Phi} \Vert \mathbf{L}^s - \Phi(\mathbf{C})^T\Phi(\mathbf{X}^s) \Vert_F^2 + \gamma \mathbb{MMD}(\mathbf{X}^s,\mathbf{X}^t),
\label{eqn:nb3}
\end{eqnarray}
where $\gamma$ is the trade-off parameter controlling the balance between the loss function and the MMD regularization term.

However, the optimization problem of Eq.~(\ref{eqn:nb3}) is a hard task because we cannot directly optimize the kernel mapping operator $\Phi$. To solve this problem, we relax the MMD distance to the following one, which is designed in our previous work of~\cite{zong2018domain}:
\begin{eqnarray}
f_G = \Vert \Phi(\mathbf{C})^T[\frac1{N_s}\Phi(\mathbf{X}^s)\mathbf{1}_s - \frac1{N_t}\Phi(\mathbf{X}^t)\mathbf{1}_t]\Vert_2^2.
\label{eqn:nb4}
\end{eqnarray}
By resorting to $f_G$, the optimization of kernel mapping operator $\Phi(\cdot)$ in minimizing MMD is then changed to the optimization problem with respect to $\Phi(\mathbf{C})$.  Thus, our transfer regression model  in Eq.~(\ref{eqn:nb4}) can be formulated as:
\begin{eqnarray}
\min_{\Phi(\mathbf{C})} \Vert \mathbf{L}^s - \Phi(\mathbf{C})^T\Phi(\mathbf{X}^s) \Vert_F^2~~~~~~~~~~~~~~~~~~~~~~~~~~ \nonumber \\+ \gamma \Vert \Phi(\mathbf{C})^T[\frac1{N_s}\Phi(\mathbf{X}^s)\mathbf{1}_s - \frac1{N_t}\Phi(\mathbf{X}^t)\mathbf{1}_t]\Vert_2^2.
\label{eqn:nb5}
\end{eqnarray}

Note that the transfer regression model in Eq.(\ref{eqn:nb5}) is still a general DA method rather than an expert CDMER one because this model can be used to deal with any unsupervised DA problem. It can be also seen that any background knowledge of micro-expressions is not considered during its construction. In fact, in recent years various cues in micro-expression samples, e.g., the color information~\cite{wang2015micro}, facial local region information~\cite{zong2018learning}, and dynamic sparsity~\cite{le2016sparsity}, have been explored to demonstrate their effectiveness in distinguishing different micro-expressions. Therefore, it would be beneficial to take these positive cues of micro-expressions into consideration when we design CDMER methods. To this end, we would like to leverage the idea of the facial local region selection in our previous work of~\cite{zong2018learning} to enhance our transfer regression model in dealing with CDMER problem.

Specifically, let $\phi(\cdot)$ be another kernel mapping operator that has the relationship with $\Phi(\cdot)$, i.e., $\Phi(\mathbf{X}) = [\phi(\mathbf{X}_1)^T,\cdots,\phi(\mathbf{X}_K)^T]^T$. Then, the regression coefficient matrix $\Phi(\mathbf{C})$ can be rewritten as $\Phi(\mathbf{C}) = [\phi(\mathbf{C}_1)^T, \cdots, \phi(\mathbf{C}_K)^T]^T$ and hence we are able to reformulate the optimization problem of Eq.(\ref{eqn:nb5}) as follows:
\begin{eqnarray}
\min_{\phi(\mathbf{C}_i)} \Vert \mathbf{L}^s - \sum_{i=1}^K \phi(\mathbf{C}_i)^T\phi(\mathbf{X}_i^s) \Vert_F^2~~~~~~~~~~~~~~~~~~~~~~~~ \nonumber \\+ \gamma \Vert \sum_{i=1}^K \phi(\mathbf{C}_i)^T[\frac1{N_s}\phi(\mathbf{X}_i^s)\mathbf{1}_s - \frac1{N_t}\phi(\mathbf{X}_i^t)\mathbf{1}_t]\Vert_2^2,
\label{eqn:nb6}
\end{eqnarray}

In order to pick out the micro-expression related local facial regions and leverage them to enhance our transfer regression model in Eq.(\ref{eqn:nb6}), we endow a non-negative weighted parameter $w_i$ to the $i^{th}$ facial block $\mathbf{X}_i^s$ and $\mathbf{X}_i^t$ to measure its specific contributions to distinguish different micro-expresisons. To achieve this goal, we use a L1-norm with respect to $\mathbf{w} = [w_1,\cdots,w_K]^T$ to serve as the regularization term for our transfer regression model and then we arrive at the following new formulation:
\begin{eqnarray}
\min_{\phi(\mathbf{C}_i),\mathbf{w}} \Vert \mathbf{L}^s - \sum_{i=1}^K w_i\phi(\mathbf{C}_i)^T\phi(\mathbf{X}_i^s) \Vert_F^2 + \lambda \Vert \mathbf{w} \Vert_1~~~~~~~~ \nonumber \\+ \gamma \Vert \sum_{i=1}^K w_i\phi(\mathbf{C}_i)^T[ \frac1{N_s}\phi(\mathbf{X}_i^s)\mathbf{1}_s - \frac1{N_t}\phi(\mathbf{X}_i^t)\mathbf{1}_t]\Vert_2^2,\nonumber\\
\textup{s.t.}~\mathbf{w} \succeq 0.~~~~~~~~~~~~~~~~~~~~~~~~~~~~~~~~~~~~~~~~~~~~~~
\end{eqnarray}

To differentiate this new transfer regression model with traditional ones, we call it \textbf{Region Selective Transfer Regression (RSTR)}. Via the kernel trick, the explicit mapping $\phi(\cdot)$ in RSTR can be effectively avoided and further optimized. According to the kernel representation theory \cite{scholkopf1998nonlinear}, we obtain that $\Phi(\mathbf{C})$ can be expressed as $\Phi(\mathbf{C})= [\Phi(\mathbf{X}^s),\Phi(\mathbf{X}^t)]\mathbf{P}$ resulting in $\phi(\mathbf{C}_i) = [\phi(\mathbf{X}^s_i), \phi(\mathbf{X}^t_i)] \mathbf{P}$, where $\mathbf{P}$ is a coefficient matrix. We can then arrive at the final proposed RSTR, which has the formulation as follows:
\begin{eqnarray}
\min_{\mathbf{P},\mathbf{w}} \Vert \mathbf{L}^s - \mathbf{P}^T\sum_{i=1}^K w_i \mathbf{K}_i^s \Vert_F^2 + \lambda \Vert \mathbf{w} \Vert_1+ \mu \Vert \mathbf{P} \Vert_1~~~ \nonumber \\+ \gamma \Vert \sum_{i=1}^K w_i\mathbf{P}^T(\frac1{N_s}\mathbf{K}_i^s\mathbf{1}_s - \frac1{N_t}\mathbf{K}_i^t\mathbf{1}_t)\Vert_2^2,\nonumber\\
\textup{s.t.}~\mathbf{w} \succeq 0.~~~~~~~~~~~~~~~~~~~~~~~~~~~~~~~~~~~~~~~~~~~
\label{eqn:nb8}
\end{eqnarray}
where $\lambda$, $\mu$, and $\gamma$ are the trade-off parameters which control the balance between the loss function of RSTR and the regularization terms, and $\mathbf{K}_i^s = \left[\begin{array}{c} \phi(\mathbf{X}_i^s)^T\\ \phi(\mathbf{X}_i^t)^T\end{array}\right]\phi(\mathbf{X}_i^s)$ and $\mathbf{K}_i^t = \left[\begin{array}{c} \phi(\mathbf{X}_i^s)^T\\ \phi(\mathbf{X}_i^t)^T\end{array}\right]\phi(\mathbf{X}_i^t)$ and can be computed by using the preset kernel functions, e.g., Linear, Polynomial, and Gaussian kernels. Note that following~\cite{zong2018learning}, we also add a L1-norm of $\mathbf{P}$, i.e., $\Vert \mathbf{P} \Vert_1 = \sum_{i=1}^{c}\Vert \mathbf{p}_i \Vert_1$, where $\mathbf{p}_i$ is the $i^{th}$ column of $\mathbf{P}$, for Eq.(\ref{eqn:nb8}) as the regularization term such that the regression coefficient matrix $\phi(\mathbf{C}_i)$ can be sparsely reconstructed by $[\phi(\mathbf{X}_i^s),\phi(\mathbf{X}_i^t)]$ and the overfitting during the optimization can be avoided.

\subsection{Optimization}
\label{opt}
The proposed RSTR can be easily solved by the alternated direction method (ADM)~\cite{zheng2014multi}, i.e., fixing one parameter and updating the other ones until convergence. More specifically, repeat the following three steps:
\subsubsection{Fix $\mathbf{w}$ and Update $\mathbf{P}$}
In this step, the optimization problem with respect to $\mathbf{P}$ can be written as
\begin{eqnarray}
\min_{\mathbf{P}} \Vert \mathbf{L}^s - \mathbf{P}^T \tilde{\mathbf{K}}^s \Vert_F^2 + \mu \Vert \mathbf{P} \Vert_1 + \gamma \Vert \mathbf{P}^T\tilde{\mathbf{k}}_{st}\Vert_2^2,
\label{eqn:nb9}
\end{eqnarray}
where $\tilde{\mathbf{K}}^s = \sum_{i=1}^K w_i \mathbf{K}_i^s$ and $\tilde{\mathbf{k}}_{st} = \sum_{i=1}^K w_i(\frac1{N_s}\mathbf{K}_i^s\mathbf{1}_s - \frac1{N_t}\mathbf{K}_i^t\mathbf{1}_t)$. In this paper, we adopt the inexact augmented Lagrangian multiplier (IALM) approach~\cite{lin2010augmented} to learn the optimal $\mathbf{P}$ in Eq.~(\ref{eqn:nb9}). More specifically, we introduce an auxiliary variable $\mathbf{Q}$ that equals $\mathbf{P}$ to convert the unconstrained optimization problem in Eq.~(\ref{eqn:nb9}) to a constrained one as follows:
\begin{eqnarray}
\min_{\mathbf{P},\mathbf{Q}} \Vert \mathbf{L}^s - \mathbf{Q}^T \tilde{\mathbf{K}}^s \Vert_F^2 + \mu \Vert \mathbf{P} \Vert_1 + \gamma \Vert \mathbf{Q}^T\tilde{\mathbf{k}}_{st}\Vert_2^2,\nonumber\\
\textup{s.t.}~\mathbf{P} = \mathbf{Q}.~~~~~~~~~~~~~~~~~~~~~~~~~~~~~~~~~~~~~~~
\label{eqn:nb10}
\end{eqnarray}
Subsequently, we can obtain the Lagrangian function of Eq.~(\ref{eqn:nb10}), which has the following formulation:
\begin{eqnarray}
L(\mathbf{P},\mathbf{Q},\mathbf{T},\kappa) = \Vert \mathbf{L}^s - \mathbf{Q}^T \tilde{\mathbf{K}}^s \Vert_F^2 + \mu \Vert \mathbf{P} \Vert_1 ~~~~~~~~~~~~~~\nonumber\\ + \gamma \Vert \mathbf{Q}^T\tilde{\mathbf{k}}_{st}\Vert_2^2 + tr(\mathbf{T}^T(\mathbf{P} - \mathbf{Q})) + \frac{\kappa}2\Vert \mathbf{P} - \mathbf{Q} \Vert_F^2,
\label{eqn:nb11}
\end{eqnarray}
where $\kappa$ is a trade-off parameter and $\mathbf{T}$ is the Lagrangian multiplier matrix. We only need to iteratively minimize the Lagrangian function in Eq.~(\ref{eqn:nb11}) with respect to its different variables and then the optimal $\mathbf{P}$ can be learned. We summarize the complete updating procedures in Algorithm~\ref{alm:nb1}.

\begin{algorithm}[t]
\caption{Complete updating rule for learning the optimal $\mathbf{P}$ in Eq. (\ref{eqn:nb9}).}
\textbf{Repeating Steps 1) to 4) until obtaining convergence:} \\
1) Fix $\mathbf{P}$, $\mathbf{T}$, and $\kappa$ and update $\mathbf{Q}$:
\begin{eqnarray}
\min_\mathbf{Q} \Vert \mathbf{L}^s - \mathbf{Q}^T \tilde{\mathbf{K}}^s \Vert_F^2 + \gamma \Vert \mathbf{Q}^T\tilde{\mathbf{k}}_{st}\Vert_2^2 + tr(\mathbf{T}^T(\mathbf{P} - \mathbf{Q})) \nonumber\\ + \frac{\kappa}2\Vert \mathbf{P} - \mathbf{Q} \Vert_F^2,~~~~~~~~~~~~~~~~~~~~~~~~~~~~~~~~~~~~~~~\nonumber
\end{eqnarray}
which has the close-form solution as follows:
\begin{eqnarray}
\mathbf{Q} = (\tilde{\mathbf{K}}^s \tilde{\mathbf{K}}^s + \sqrt{\gamma}\tilde{\mathbf{k}}_{st}\tilde{\mathbf{k}}_{st}^T+ \frac{\kappa}2 \mathbf{I})^{-1} (\tilde{\mathbf{K}}^s {\mathbf{L}^s}^T + \frac{\mathbf{T}+\kappa \mathbf{P}}2), \nonumber
\end{eqnarray}
where $\mathbf{I}$ is an identity matrix.

2) Fix $\mathbf{Q}$, $\mathbf{T}$ and $\kappa$ and update $\mathbf{P}$:
\begin{eqnarray}
\min_\mathbf{P} \frac{\mu}{\kappa} \Vert \mathbf{P} \Vert_1 + \frac12 \Vert \mathbf{P} - (\mathbf{Q} - \frac{\mathbf{T}}{\kappa}) \Vert_F^2. \nonumber
\end{eqnarray}
The solution of the above optimization problem is $S_{\frac{\mu}{\kappa}}[\mathbf{Q} - \frac{\mathbf{T}}{\kappa}]$, where $S_{\zeta}[A]$ is the soft thresholding operator and is defined as:
\begin{eqnarray}
S_{\zeta}[A] = \left\{\begin{array}{l c}
A - \zeta,~~~\textup{if } A > \zeta;\nonumber\\
A + \zeta,~~~\textup{if } A < -\zeta;\nonumber\\
0,~~~~~~~~~\textup{otherwise}.\nonumber
\end{array}\right.
\end{eqnarray}

3) Update $\mathbf{T}$ and $\kappa$:
\begin{eqnarray}
\mathbf{T} = \mathbf{T} + \kappa(\mathbf{P} - \mathbf{Q}),\kappa = \min(\rho\kappa,\kappa_{max}),
\nonumber
\end{eqnarray}
where $\kappa_{max}$ is a preset maximal value for $\kappa$ and $\rho$ is a scaled parameter and can be set as a value greater than 1.

4) Check convergence:
\begin{eqnarray}
\Vert \mathbf{P} - \mathbf{Q} \Vert_\infty < \epsilon,\nonumber
\end{eqnarray}
where $\infty$ norm means the maximal value of the element in a matrix and $\epsilon$ is the machine epsilon.
\label{alm:nb1}
\end{algorithm}

\subsubsection{Fix $\mathbf{P}$ and Update $\mathbf{w}$}

\begin{eqnarray}
\min_{\mathbf{w}} \Vert \mathbf{z} - \mathbf{A}\mathbf{w} \Vert_2^2 + \gamma \Vert \mathbf{B}\mathbf{w}\Vert_2^2 + \lambda \Vert \mathbf{w} \Vert_1,\nonumber\\
\textup{s.t.}~\mathbf{w} \succeq 0.~~~~~~~~~~~~~~~~~~~~~~~~~~~~~~~
\label{eqn:nb12}
\end{eqnarray}
Herein, $\mathbf{z}$ is a vector composed by concatenating the columns of $\mathbf{L}^s$ one by one, and the $i^{th}$ columns of $\mathbf{A}$ and $\mathbf{B}$ are a vector by vectorizing their corresponding matrices $\mathbf{P}^T\mathbf{K}_i^s$, and $\mathbf{P}^T(\frac1{N_s}\mathbf{K}_i^s\mathbf{1}_s - \frac1{N_t}\mathbf{K}_i^t\mathbf{1}_t)$, respectively. Eq.~(\ref{eqn:nb10}) can be further rewritten as the following standard Lasso problem:
\begin{eqnarray}
\min_{\mathbf{w}} \Vert \mathbf{y} - \mathbf{D}\mathbf{w} \Vert_2^2 + \lambda \Vert \mathbf{w} \Vert_1,\nonumber\\
\textup{s.t.}~\mathbf{w} \succeq 0.~~~~~~~~~~~~~~~~~~
\end{eqnarray}
where $\mathbf{y} = \left[\begin{array}{c} \mathbf{z} \\ \mathbf{0} \end{array}\right]$ and $\mathbf{D} = \left[\begin{array}{c} \mathbf{A} \\ \mathbf{B} \end{array}\right]$. In our optimization, we use the SLEP package~\cite{liu2009slep} to learn the optimal $\mathbf{w}$.

\subsubsection{Check Convergence}

Check whether the preset maximal iteration steps are reached or the value of the objective function in Eq.~(\ref{eqn:nb8}) is less than the preset threshold.

\label{optimization}

\subsection{RSTR for Solving CDMER Problem}

Once the optimal parameters of RSTR are learned based on the labeled source and unlabeled target micro-expression samples using the optimization method given in Section~\ref{opt}, we can easily predict the micro-expression categories of the testing samples from target database. More specifically, suppose the learned optimal parameters of RSTR are $\hat\mathbf{P}$ and $\hat\mathbf{w}$. Then, given a testing target micro-expression sample $\mathbf{x}^{te}$ we are able to estimate its micro-expression label vector by solving the following optimization problem:
\begin{eqnarray}
\min_{\mathbf{l}^{te}} \Vert \mathbf{l}^{te} - \hat\mathbf{P}^T \sum_{i=1}^M \hat\omega_i \mathbf{k}_i^{te} \Vert_2^2,\nonumber \\
\textup{s.t.}~\mathbf{l}^{te} \succeq 0,~\mathbf{1}^T\mathbf{l}^{te} = 1,~~~~
\label{eqn:nb12}
\end{eqnarray}
where $\hat\omega_i$ is the $i^{th}$ element of $\hat\mathbf{w}$ and $\mathbf{k}_i^{te} = [\phi(\mathbf{X}_i^s), \phi(\mathbf{X}_i^t)]^T\phi(\mathbf{x}_i^{te})$ is the kernel matrix and can be computed by the kernel function chosen in training stage. Finally, the micro-expression category of this testing sample is assigned as the one whose corresponding value in $\mathbf{l}^{te}$ is maximum.

\section{Evaluation Results}
\label{experiment}
\subsection{Implementation Details}
We conduct the benchmark CDMER experiments under the designed protocol described in Section~\ref{benchmark}. For the evaluated methods, we implement them using the original source codes provided by the authors. To offer a fair comparison, for the experiments of different micro-expression features, the parameters of each spatiotemporal descriptors are fixed throughout the experiments, while for the experiments of DA, we follow the strategy used in current mainstream unsupervised DA evaluation experiments that is reporting the best result (in term of $mean$ $F1$-$score$ in our CDMER benchmark experiments) corresponding to the optimal parameter setting for a DA method~\cite{ding2014latent,al2014supervised,long2014transfer,long2015domain,zong2017learning}.
Subsequently, we show the detailed parameter settings of both micro-expression features and DA methods used for our benchmark experiments.

\subsubsection{Parameter Setting for DA Methods}

In this section, we give the parameter searching space for different DA methods in the CDMER evaluation experiments.

\textbf{SVM}~\cite{chang2011libsvm}: To offer a fair comparison, we use linear kernel for SVM throughout the evaluation experiments. As to the penalized coefficient $C$, we fix it at $C=1$. Note that for the experiments of all the DA methods, we all use the linear SVM with $C=1$ to serve as the classifier (if needed).

\textbf{IW-SVM}~\cite{hassan2013acoustic}: In our experiments, we choose uLSIF~\cite{kanamori2009least}, which has shown its excellent performance in CDMER~\cite{zong2017learning,zong2018domain}, to learn the importance weights for IW-SVM. Following the suggestion of~\cite{zong2017learning,zong2018domain}, we search the trade-off parameter $\lambda$ for uLSIF from a parameter space [1:1:100]$\times t$~($t$ = 1, 10, 100, 1000, 10000, 100000).


\textbf{TCA}~\cite{pan2011domain}, \textbf{GFK}~\cite{gong2012geodesic}, and \textbf{SA}~\cite{fernando2013unsupervised}: Among the experiments of these three methods, principal component analysis (PCA)~\cite{wold1987principal} is used to construct the subspace for GFK and SA. For all of them, we search the optimal dimension $k$ (the number of eigenvectors for composing the projection matrix) by trying all possible dimensions, i.e., searching $k \in [1,2,\cdots,k_{max}]$.

\textbf{STM}~\cite{chu2013selective,chu2017selective}: STM is originally a binary classification model. In our experiments, we extend it to a multi-class version by using one-against-rest strategy. Similar with SVM used in the benchmark evaluation, its penalized coefficient is set as $C=1$. As to its second trade-off parameter $\lambda$, which is used to balance the KMM regularization term with the SVM objective function, the searching space is set as $[0.001:0.001:0.009, 0.01:0.01:0.09, 0.1:0.1:1, 2:1:100, 1000, 10000]$.

\textbf{TKL}~\cite{long2015domain}: According to the work of~\cite{long2015domain}, TKL has one important parameter called the eigenspectrum damping factor $\zeta$. In the evaluation experiments, we determine its optimal value by searching from the parameter space $[0.1:0.1:5]$.

\textbf{TSRG}~\cite{zong2017learning},~\textbf{DRFS-T}~\cite{zong2018domain}, and~\textbf{DRLS}~\cite{zong2018domain}: TSRG, DRFS-T, and DRLS have two important trade-off parameters, i.e., $\lambda$ and $\mu$. Following the works of~\cite{zong2017learning,zong2018domain}, the optimal values of these two parameters are determined by searching from $[0.001, 0.01, 0.1, 1, 10, 100, 1000]$ for $\lambda$ and $[0.001:0.001:0.009, 0.01:0.01:0.09, 0.1:0.1:1, 2:1:10]$ for $\mu$.

\textbf{RSTR}: The proposed RSTR have three trade-off parameters including $\lambda$, $\mu$, and $\tau$ which control the balance between the loss function and the regularized terms. We search their optimal values from the preset parameter spaces, i.e., $\lambda \in [0.1, 1, 10, 100, 1000, 10000]$, $\mu \in [0.1:0.1:5]$, and $\tau \in [0.01:0.01:0.1]$.

\subsubsection{Parameter Setting for Micro-Expression Features}
We set the parameters for the evaluated spatiotemporal descriptors as follows:

\textbf{LBP-TOP}~\cite{zhao2007dynamic}: LBP-TOP has two important parameters. One is the neighboring radius $R$ and the other is the number of the neighboring points $P$. In the evaluation experiments, we set the values of these two parameters as the ones of $R\in\{1,3\}$ and $P\in\{4,8\}$, respectively, and hence we report FOUR experimental results of LBP-TOP with different parameter setting, i.e., LBP-TOP$_{(R1P4)}$, LBP-TOP$_{(R1P8)}$, LBP-TOP$_{(R3P4)}$, and LBP-TOP$_{(R3P8)}$. In addition, the uniform pattern is used for LBP coding. We use Hong et al.'s fast LBP-TOP~\cite{hong2016lbp} source code to implement the LBP-TOP feature extraction.

\textbf{LBP-SIP}~\cite{wang2014lbp}: Only one parameter needs to be set for LBP-SIP, i.e., the neighboring radius $R$. Similar to LBP-TOP, we set $R$ as 1 and 3 for LBP-SIP, respectively, and report the experimental results of LBP-SIP$_{(R1)}$ and LBP-SIP$_{(R3)}$. In the experiments, LBP-SIP is implemented by ourselves.

\textbf{LPQ-TOP}~\cite{paivarinta2011volume}: For LPQ-TOP, we set its parameters following the suggestion of~\cite{paivarinta2011volume}. Specifically, the size of the local window in each dimension is set as the default one ([5, 5, 5]). The parameters $[\rho_s, \rho_t]$ for a correlation model used in LPQ-TOP are set as [0.1, 0.1] and [0, 0] (without correlation), respectively. The results of LPQ-TOP$_{decorr = 0.1}$ and LPQ-TOP$_{decorr = 0}$ are reported in the evaluation experiments. The source code of LPQ-TOP is publicly available at http://www.cse.oulu.fi/Downloads/LPQMatlab/.

\textbf{HOG-TOP} and \textbf{HIGO-TOP}~\cite{li2017towards}: HOG-TOP and HIGO-TOP both have one important parameter, i.e., the number of bins $p$ to be set, which controls the dimensional histogram of image gradient orientations for three orthogonal planes. In the evaluation experiments, we fix $p$ at 4 and 8 for HOG-TOP and HIGO-TOP and they are denoted by HOG-TOP$_{p = 4}$, HOG-TOP$_{p = 8}$, HIGO-TOP$_{p = 4}$ and HIGO-TOP$_{p = 8}$, respectively.

\textbf{C3D}~\cite{tran2015learning}: In the evaluation experiments, we choose the publicly released C3D pretrained on Sports-1M~\cite{karpathy2014large} and UCF101~\cite{soomro2012ucf101} as the feature extractor and use the last two fully connected layers to serve as the micro-expression features, which are denoted by C3D-FC1 (Sports-1M) and C3D-FC2 (Sports-1M), respectively.

\subsection{Results and Discussions}

\subsubsection{Results at A Glance}

In this section, we report the benchmark results of the evaluated methods including various DA methods and micro-expression features. All the experimental results are depicted in Tables~\ref{tab:tab3},~\ref{tab:tab4},~\ref{tab:tab5}, and~\ref{tab:tab6}. Before deeply comparing and analyzing the results obtained by different DA methods and micro-expressions, we would like to probe into the CDMER tasks based on the obtained results. We calculate the average results of all the methods in each experiment and all the experiments for each method, which are given in the last line and last column of these tables. From the comparison between them, we are able to reach the following conclusions.

Firstly, we observe that for our designed benchmark, the second type of experiments (TYPE-II) are significantly more difficult than TYPE-I, which can be clearly revealed by the remarkable differences between the average results of each method in these two types of experiments. For example, the baseline method, GFK (a well-performing representative of DA methods), and LPQ-TOP$_{decorr=0.1}$ (a well-performing representative of micro-expression features) achieve the average $mean$ $F1$-$score$ / $Accuracy$ of 0.6003 / 61.62\%, 0.7223 / 72.44\%, and 0.6157 / 63.79\% in the first type of experiments, which are much higher than their achieved results (0.4112 / 45.55\%, 0.5161 / 54.31\%, and 0.3236 / 38.51) in the second type of experiments. In fact, the differences in difficulty between the TYPE-I and TYPE-II experiments stand to reason. The three datasets (HS, VIS, and NIR) used in TYPE-I involve the same subjects, stimulus materials, and recording environments and just different cameras, which results in the relatively small dataset difference. However, another dataset used in TYPE-II experiments, CASME~II, corresponds to substantially different subjects, stimulus material, and recording environments compared with SMIC (HS, VIS, and NIR).

Secondly, it should be pointed out that the class imbalanced problem existing in the source or target database remarkably degrades the performance of either DA method or micro-expression features in dealing with the CDMER tasks. For example, in the cases with SMIC (NIR) as the target database, i.e., Exp.11, Exp.3, and Exp.5, we can observe that the average performance in terms of $mean$ $F1$-$score$ and $Accuracy$ of all the DA methods can reach 0.6881 and 70.42\% in Exp.5 whose the source database, SMIC (VIS), is relatively class-balanced. These two metrics drop to 0.6782 and 67.86\% in Exp.3 and 0.5011 and 51.39\% in Exp.11, where the source databases of Exp.3 and Exp.11 are SMIC (HS) and CASME~II, respectively and very class-imbalanced. Similarly, the performance of all the methods is also affected by the class-imbalanced target database. As shown in Exp.6, Exp.3, and Exp.12 whose source database is fixed, i.e., SMIC (NIR), it can be seen that with class-imbalanced database as target one, the average $mean$ $F1$-$score$ / $Accuracy$ decrease from the level of 0.7272 / 73.88\% (Exp.6: class-balanced) to 0.5668 / 57.26\% (Exp.4: class-imbalanced) and 0.3928 / 41.96\% (Exp.4: class-imbalanced), respectively.

Thirdly, we can also observe that the heterogeneous problem existing between source and target databases raises the level of difficulty of the CDMER tasks. It is known that the samples in SMIC (NIR) are recorded by a near-infrared camera, whose image-quality is considerably different from the samples recorded by high-speed camera (used in CASME~II and SMIC (HS)) and visual camera (used in SMIC (VIS)). Therefore, it is intuitive that the CDMER tasks involving SMIC (NIR) would be more difficult than others. In order to check this point, we first fix the source database as SMIC (HS) and observe two experiments, i.e., Exp.1 and Exp.3, where the target database in Exp.1 is SMIC (VIS) and Exp.3 corresponds to SMIC (NIR). It can be seen that the average results among all the DA methods and micro-expression features (0.8405 / 84.12\% and 0.6459 / 67.52\%) in Exp.1 (homogeneous) are significantly higher than the results (0.6782 / 67.86\% and 0.4420 / 49.65\%) in Exp.3 (heterogeneous). We further check the opposite case if the heterogeneous image-quality samples exist in the source database. We observe Exp.2 and Exp.4, whose target databases are the same, i.e., SMIC (HS) and source databases are different, i.e., SMIC (VIS) v.s. SMIC (NIR). From the results, we notice the performance difference between the Exp.2 (homogeneous case) and Exp.4 (heterogeneous case). Specifically, the average performance achieved by DA methods and micro-expressions are 0.5764 / 57.60\% and 0.4318 / 45.35\% in Exp.2 (homogeneous case) v.s. 0.5668 / 57.26\% and 0.3616 / 39.18\% in Exp.4 (heterogeneous case).

\begin{table*}[!t]
\scriptsize
\renewcommand{\arraystretch}{1.2}
\caption{Experimental results (mean F1-score / Accuracy) of various domain adaptation methods for CDMER, where the source and target databases are two subsets of SMIC (HS, VIS, and NIR). The micro-expression categories (3 classes) are $Negative$, $Positive$, and $Surprise$. The best results in each experiment are highlighted in bold.}
\label{tab:tab3}
\centering
\begin{tabular}{|l|c|c|c|c|c|c|c|}
\hline
DA Method & Exp.1: H $\rightarrow$ V & Exp.2: V $\rightarrow$ H & Exp.3: H $\rightarrow$ N & Exp.4: N $\rightarrow$ H & Exp.5: V $\rightarrow$ N & Exp.6: N $\rightarrow$ V & Average\\ \hline\hline
Baseline & 0.8002 / 80.28 & 0.5421 / 54.27 & 0.5455 / 53.52 & 0.4878 / 54.88 & 0.6186 / 63.38 & 0.6078 / 63.38 & 0.6003 / 61.62\\ \hline
IW-SVM~\cite{hassan2013acoustic} & 0.8868 / \textbf{88.73} & 0.5852 / 58.54 & \textbf{0.7469 / 74.65} & 0.5427 / 54.27 & 0.6620 / 69.01 & 0.7228 / 73.24 & 0.6911 / 68.07 \\ \hline
TCA~\cite{pan2011domain} & 0.8269 / 83.10 & 0.5477 / 54.88 & 0.5828 / 59.15 & 0.5443 / 57.32 & 0.5810 / 61.97 & 0.6598 / 67.61 & 0.6238 / 64.01\\ \hline
GFK~\cite{gong2012geodesic} & 0.8448 / 84.51 & 0.5957 / 59.15 & 0.6977 / 70.42 & \textbf{0.6197 / 62.80} & \textbf{0.7619 / 76.06} & 0.8142 / 81.69 & 0.7223 / 72.44 \\ \hline
SA~\cite{fernando2013unsupervised} & 0.8037 / 80.28 & 0.5955 / 59.15 & 0.7465 / \textbf{74.65} & 0.5644 / 56.10 & 0.7004 / 71.83 & 0.7394 / 74.65 & 0.6917 / 69.44\\ \hline
STM~\cite{chu2013selective,chu2017selective} & 0.8253 / 83.10 & 0.5059 / 51.22 & 0.6628 / 66.20 & 0.5351 / 56.10 & 0.6427 / 67.61 & 0.6922 / 70.42 & 0.6440 / 65.78\\ \hline
TKL~\cite{long2015domain} & 0.7742 / 77.46 & 0.5738 / 57.32 & 0.7051 / 70.42 & 0.6116 / 62.20 & 0.7558 / \textbf{76.06} & 0.7579 / 76.06 & 0.6964 / 69.92\\ \hline
TSRG~\cite{zong2017learning} & \textbf{0.8869 / 88.73} & 0.5652 / 56.71 & 0.6484 / 64.79 & 0.5770 / 57.93 & 0.7056 / 70.42 & 0.8116 / 81.69 & 0.6991 / 70.05\\ \hline
DRFS-T~\cite{zong2018domain} & 0.8643 / 85.92 & 0.5767 / 57.32 & 0.7179 / 71.83 & 0.6163 / 61.59 & 0.7286 / 73.24 & 0.7732 / 77.46 & 0.7128 / 71.23\\ \hline
DRLS~\cite{zong2018domain} & 0.8604 / 85.92 & 0.6120 / 60.98 & 0.6599 / 66.20 & 0.5599 / 55.49 & 0.6620 / 69.01 & 0.5771 / 61.97 & 0.6552 / 66.60 \\ \hline
RSTR & 0.8721 / 87.32 & \textbf{0.6401 / 64.02} & 0.7466 / \textbf{74.65} & 0.5765 / 57.32 & 0.7506 / \textbf{76.06} & \textbf{0.8428 / 84.51} & \textbf{0.7381 / 73.98} \\ \hline\hline
Average & 0.8405 / 84.12 & 0.5764 / 57.60 & 0.6782 / 67.86 & 0.5668 / 57.26 & 0.6881 / 70.42 & 0.7272 / 73.88 & - \\ \hline
\end{tabular}
\end{table*}

\begin{table*}[!t]
\scriptsize
\renewcommand{\arraystretch}{1.2}
\caption{Experimental results (mean F1-score / Accuracy) of various domain adaptation methods for CDMER, where the source and target databases are CASME~II or one subset of SMIC (HS, VIS, and NIR). The micro-expression categories (3 classes) are $Negative$, $Positive$, and $Surprise$. The best results in each experiment are highlighted in bold.}
\label{tab:tab4}
\centering
\begin{tabular}{|l|c|c|c|c|c|c|c|}
\hline
DA Method & Exp.7: C $\rightarrow$ H & Exp.8: H $\rightarrow$ C & Exp.9: C $\rightarrow$ V & Exp.10: V $\rightarrow$ C & Exp.11: C $\rightarrow$ N & Exp.12: N $\rightarrow$ C & Average\\ \hline\hline
Baseline & 0.3697 / 45.12 & 0.3245 / 48.46 & 0.4701 / 50.70 & 0.5367 / 53.08 & 0.5295 / 52.11 & 0.2368 / 23.85 & 0.4112 / 45.55\\ \hline
IW-SVM~\cite{hassan2013acoustic} & 0.3541 / 41.46 & 0.5829 / 62.31 & 0.5778 / 59.15 & 0.5537 / 54.62 & 0.5117 / 50.70 & 0.3456 / 36.15 & 0.4876 / 50.73\\ \hline
TCA~\cite{pan2011domain} & 0.4637 / 46.34 & 0.4870 / 53.08 & \textbf{0.6834 / 69.01} & 0.5789 / 59.23 & 0.4992 / 50.70 & 0.3937 / 42.31 & 0.5177 / 53.45\\ \hline
GFK~\cite{gong2012geodesic} & 0.4126 / 46.95 & 0.4776 / 50.77 & 0.6361 / 66.20 & 0.6056 / 61.50 & 0.5180 / 53.52 & 0.4469 / 46.92 & 0.5161 / 54.31\\ \hline
SA~\cite{fernando2013unsupervised} & 0.4302 / 47.56 & 0.5447 / 62.31 & 0.5939 / 59.15 & 0.5243 / 51.54 & 0.4738 / 47.89 & 0.3592 / 36.92 & 0.4877 / 50.90\\ \hline
STM~\cite{chu2013selective,chu2017selective} & 0.3640 / 43.90 & \textbf{0.6115 / 63.85} & 0.4051 / 52.11 & 0.2715 / 30.00 & 0.3523 / 42.25 & 0.3850 / 41.54 & 0.3982 / 45.61\\ \hline
TKL~\cite{long2015domain} & 0.3829 / 44.51 & 0.4661 / 54.62 & 0.6042 / 60.56 & 0.5378 / 53.08 & 0.5392 / 54.93 & 0.4248 / 43.85 & 0.4925 / 51.93\\ \hline
TSRG~\cite{zong2017learning} & 0.5042 / 51.83 & 0.5171 / 60.77 & 0.5935 / 59.15 & 0.6208 / 63.08 & \textbf{0.5624 / 56.34} & 0.4105 / 46.15 & 0.5348 / 56.22 \\ \hline
DRFS-T~\cite{zong2018domain} & 0.4524 / 46.95 & 0.5460 / 60.00 & 0.6217 / 63.38 & 0.6762 / 68.46 & 0.5369 / \textbf{56.34} & 0.4653 / \textbf{50.77} & 0.5498 / 57.65\\ \hline
DRLS~\cite{zong2018domain} & 0.4924 / 53.05 & 0.5267 / 59.23 & 0.5757 / 57.75 & 0.5942 / 60.00 & 0.4885 / 49.83 & 0.3838 / 42.37 & 0.5102 / 53.71\\ \hline
RSTR & \textbf{0.5297 / 54.27} & 0.5622 / 60.77 & 0.5882 / 59.15 & \textbf{0.7021 / 70.77} & 0.5009 / 50.70 & \textbf{0.4693 / 50.77} & \textbf{0.5587 / 57.74} \\ \hline\hline
Average & 0.4323 / 47.45 & 0.5133 / 57.83 & 0.5772 / 59.66 & 0.5638 / 56.85 & 0.5011 / 51.39 & 0.3928 / 41.96 & - \\ \hline
\end{tabular}
\end{table*}

\begin{table*}[!t]
\scriptsize
\renewcommand{\arraystretch}{1.2}
\caption{Experimental results (mean F1-score / Accuracy) of various domain adaptation methods for CDMER, where the source and target databases are two subsets of SMIC (HS, VIS, and NIR). The micro-expression categories (3 classes) are $Negative$, $Positive$, and $Surprise$. The best results in each experiment are highlighted in bold.}
\label{tab:tab5}
\centering
\begin{tabular}{|l|c|c|c|c|c|c|c|}
\hline
Spatiotemporal Descriptors & Exp.1: H $\rightarrow$ V & Exp.2: V $\rightarrow$ H & Exp.3: H $\rightarrow$ N & Exp.4: N $\rightarrow$ H & Exp.5: V $\rightarrow$ N & Exp.6: N $\rightarrow$ V & Average\\ \hline\hline
Baseline & 0.8002 / 80.28 & 0.5421 / 54.27 & 0.5455 / 53.52 & \textbf{0.4878 / 54.88} & 0.6186 / 63.38 & 0.6078 / 63.38 & 0.6003 / 61.62\\ \hline
LBP-TOP$_{(R1P4)}$~\cite{zhao2007dynamic} & 0.7185 / 71.83 & 0.3366 / 40.24 & 0.4969 / 49.30 & 0.3457 / 40.24 & 0.5480 / 57.75 & 0.5085 / 59.15 & 0.4924 / 53.32 \\ \hline
LBP-TOP$_{(R1P8)}$~\cite{zhao2007dynamic} & 0.8561 / 85.92 & 0.5329 / 53.66 & 0.5164 / 57.75 & 0.3246 / 35.37 & 0.5124 / 57.75 & 0.4481 / 50.70 & 0.5318 / 56.86 \\ \hline
LBP-TOP$_{(R3P4)}$~\cite{zhao2007dynamic} & 0.4656 / 49.30 & 0.4122 / 45.12 & 0.3682 / 40.85 & 0.3396 / 40.85 & 0.5069 / 59.15 & 0.5144 / 60.56 & 0.4345 / 49.31 \\ \hline
LBP-SIP$_{(R1)}$~\cite{wang2014lbp} & 0.6290 / 63.38 & 0.3447 / 40.85 & 0.3249 / 33.80 & 0.3490 / 42.07 & 0.5477 / 60.56 & 0.5509 / 60.56 & 0.4577 / 50.20 \\ \hline
LBP-SIP$_{(R3)}$~\cite{wang2014lbp} & 0.8574 / 85.92 & 0.4886 / 50.00 & 0.4977 / 54.93 & 0.4038 / 42.68 & 0.5444 / 59.15 & 0.3994 / 46.48 & 0.5319 / 56.53 \\ \hline
LPQ-TOP$_{(decorr = 0.1)}$~\cite{paivarinta2011volume} & \textbf{0.9455 / 94.37} & 0.5523 / 54.88 & 0.5456 / 61.97 & 0.4729 / 47.56 & 0.5416 / 57.75 & 0.6365 / 66.20 & 0.6157 / 63.79 \\ \hline
LPQ-TOP$_{(decorr = 0)}$~\cite{paivarinta2011volume} & 0.7711 / 77.46 & 0.4726 / 48.78 & 0.6771 / 67.61 & 0.4701 / 48.17 & \textbf{0.7076 / 71.83} & \textbf{0.6963 / 70.42} & \textbf{0.6325 / 64.05} \\ \hline
HOG-TOP$_{(p=4)}$~\cite{li2017towards} & 0.7068 / 71.83 & \textbf{0.5649 / 57.32} & \textbf{0.6977 / 70.42} & 0.2830 / 29.27 & 0.4569 / 49.30 & 0.3218 / 36.62 & 0.4554 / 48.47 \\ \hline
HOG-TOP$_{(p=8)}$~\cite{li2017towards} & 0.7364 / 74.65 & 0.5526 / 56.10 & 0.3990 / 46.48 & 0.2941 / 32.32 & 0.4137 / 46.48 & 0.3245 / 38.03 & 0.4453 / 49.01 \\ \hline
HIGO-TOP$_{(p=4)}$~\cite{li2017towards} & 0.7933 / 80.28 & 0.4775 / 50.61 & 0.4023 / 47.89 & 0.3445 / 35.98 & 0.5000 / 53.52 & 0.3747 / 40.85 & 0.4821 / 51.52 \\ \hline
HIGO-TOP$_{(p=8)}$~\cite{li2017towards} & 0.8445 / 84.51 & 0.5186 / 53.66 & 0.4793 / 54.93 & 0.4322 / 43.90 & 0.5054 / 54.93 & 0.4056 / 46.48 & 0.5309 / 56.40\\ \hline
C3D-FC1 (Sports1M)~\cite{tran2015learning} & 0.1577 / 30.99 & 0.2188 / 23.78 & 0.1667 / 30.99 & 0.3119 / 34.15 & 0.3802 / 49.30 & 0.3032 / 36.62 & 0.2564 / 34.31 \\ \hline
C3D-FC2 (Sports1M)~\cite{tran2015learning} & 0.2555 / 36.62 & 0.2974 / 29.27 & 0.2804 / 33.80 & 0.3239 / 36.59 & 0.4518 / 47.89 & 0.3620 / 38.03 & 0.3285 / 37.03 \\ \hline
C3D-FC1 (UCF101)~\cite{tran2015learning} & 0.3803 / 46.48 & 0.3134 / 34.76 & 0.3697 / 47.89 & 0.3440 / 34.76 & 0.3916 / 47.89 & 0.2433 / 29.58 & 0.3404 / 40.23 \\ \hline
C3D-FC2 (UCF101)~\cite{tran2015learning} & 0.4162 / 46.48 & 0.2842 / 32.32 & 0.3053 / 42.25 & 0.2531 / 28.05 & 0.3937 / 47.89 & 0.2489 / 32.39 & 0.3169 / 38.23 \\ \hline\hline
Average & 0.6459 / 67.52 & 0.4318 / 45.35 & 0.4420 / 49.65 & 0.3613 / 39.18 & 0.5013 / 55.28 & 0.4341 / 48.50 & - \\ \hline
\end{tabular}
\end{table*}

\begin{table*}[!t]
\scriptsize
\renewcommand{\arraystretch}{1.2}
\caption{Experimental results (mean F1-score / Accuracy) of various domain adaptation methods for CDMER, where the source and target databases are CASME~II or one subset subset of SMIC (HS, VIS, and NIR). The micro-expression categories (3 classes) are $Negative$, $Positive$, and $Surprise$. The best results in each experiment are highlighted in bold.}
\label{tab:tab6}
\centering
\begin{tabular}{|l|c|c|c|c|c|c|c|}
\hline
Spatiotemporal Descriptors & Exp.7: C $\rightarrow$ H & Exp.8: H $\rightarrow$ C & Exp.9: C $\rightarrow$ V & Exp.10: V $\rightarrow$ C & Exp.11: C $\rightarrow$ N & Exp.12: N $\rightarrow$ C & Average\\ \hline\hline
Baseline & 0.3697 / \textbf{45.12} & 0.3245 / 48.46 & 0.4701 / 50.70 & 0.5367 / 53.08 & \textbf{0.5295 / 52.11} & 0.2368 / 23.85 & 0.4112 / 45.55\\ \hline
LBP-TOP$_{(R1P4)}$~\cite{zhao2007dynamic} & 0.3358 / 44.51 & 0.3260 / 47.69 & 0.2111 / 35.21 & 0.1902 / 26.92 & 0.3810 / 43.66 & 0.2492 / 26.92 & 0.2823 / 37.49 \\ \hline
LBP-TOP$_{(R1P8)}$~\cite{zhao2007dynamic} & 0.3680 / 43.90 & 0.3339 / 54.62 & 0.4624 / 49.30 & 0.5880 / 57.69 & 0.3000 / 33.80 & 0.1927 / 23.08 & 0.3742 / 43.73 \\ \hline
LBP-TOP$_{(R3P4)}$~\cite{zhao2007dynamic} & 0.3117 / 43.90 & 0.3436 / 44.62 & 0.2723 / 39.44 & 0.2356 / 28.46 & 0.3818 / 49.30 & 0.2332 / 25.38 & 0.2964 / 38.52 \\ \hline
LBP-SIP$_{(R1)}$~\cite{wang2014lbp} & 0.3580 / \textbf{45.12} & 0.3039 / 44.62 & 0.2537 / 38.03 & 0.1991 / 26.92 & 0.3610 / 46.48 & 0.2194 / 26.92 & 0.2825 / 38.02 \\ \hline
LBP-SIP$_{(R3)}$~\cite{wang2014lbp} & 0.3772 / 42.68 & 0.3742 / \textbf{56.15} & \textbf{0.5846 / 59.15} & \textbf{0.6065 / 60.00} & 0.3469 / 35.21 & 0.2790 / 27.69 & \textbf{0.4279 / 46.81} \\ \hline
LPQ-TOP$_{(decorr = 0.1)}$~\cite{paivarinta2011volume} & 0.3060 / 42.07 & 0.3852 / 48.46 & 0.2525 / 33.80 & 0.4866 / 47.69 & 0.3020 / 35.21 & 0.2094 / 23.85 & 0.3236 / 38.51 \\ \hline
LPQ-TOP$_{(decorr = 0)}$~\cite{paivarinta2011volume} & 0.2368 / 43.90 & 0.2890 / 51.54 & 0.2531 / 38.03 & 0.3947 / 40.77 & 0.2369 / 35.21 & \textbf{0.4008} / 41.54 & 0.3019 / 41.83 \\ \hline
HOG-TOP$_{(p=4)}$~\cite{li2017towards} & 0.3156 / 34.76 & 0.3502 / 47.69 & 0.3266 / 35.21 & 0.4658 / 46.92 & 0.3219 / 35.21 & 0.2163 / 27.46 & 0.3327 / 37.91 \\ \hline
HOG-TOP$_{(p=8)}$~\cite{li2017towards} & \textbf{0.3992} / 43.90 & \textbf{0.4154} / 52.31 & 0.4403 / 45.07 & 0.4678 / 47.69 & 0.4107 / 40.85 & 0.1390 / 20.77 & 0.3787 / 41.77 \\ \hline
HIGO-TOP$_{(p=4)}$~\cite{li2017towards} & 0.2945 / 39.63 & 0.3420 / 53.85 & 0.3236 / 40.85 & 0.5590 / 55.38 & 0.2887 / 29.58 & 0.2668 / 31.54 & 0.3458 / 41.81 \\ \hline
HIGO-TOP$_{(p=8)}$~\cite{li2017towards} & 0.2978 / 41.46 & 0.3609 / 50.00 & 0.3679 / 43.66 & 0.5699 / 54.62 & 0.3395 / 33.80 & 0.1743 / 22.31 & 0.3517 / 40.98 \\ \hline
C3D-FC1 (Sports1M)~\cite{tran2015learning} & 0.1994 / 42.68 & 0.2394 / 56.15 & 0.1631 / 32.39 & 0.1075 / 19.23 & 0.1631 / 32.39 & 0.2397 / \textbf{56.15} & 0.1854 / 39.83 \\ \hline
C3D-FC2 (Sports1M)~\cite{tran2015learning} & 0.1994 / 42.68 & 0.1317 / 24.62 & 0.1631 / 32.39 & 0.1075 / 19.23 & 0.1631 / 32.39 & 0.2397 / \textbf{56.15} & 0.1674 / 34.58 \\ \hline
C3D-FC1 (UCF101)~\cite{tran2015learning} & 0.1581 / 31.10 & 0.1075 / 19.23 & 0.1886 / 39.44 & 0.1075 / 19.23 & 0.1886 / 39.44 & 0.2397 / \textbf{56.15} & 0.1650 / 34.10 \\ \hline
C3D-FC2 (UCF101)~\cite{tran2015learning} & 0.1994 / 42.68 & 0.1705 / 19.23 & 0.1631 / 32.39 & 0.1075 / 19.23 & 0.1631 / 32.39 & 0.1075 / 19.23 & 0.1414 / 27.53 \\ \hline\hline
Average & 0.2954 / 41.88 & 0.2959 / 44.95 & 0.3060 / 40.32 & 0.3581 / 38.94 & 0.3049 / 37.94 & 0.2277 / 31.80 & - \\ \hline
\end{tabular}
\end{table*}

\subsubsection{Results for CDMER Experiments of Using DA Methods}
Tables~\ref{tab:tab3} and~\ref{tab:tab4} show the experimental results of different DA methods corresponding to the TYPE-I and TYPE-II experiments, respectively. From these two tables, it can be seen that in both two types of designed CDMER experiments, nearly all the DA methods can achieve promisingly better results in terms of both mean F1-score and Accuracy than the baseline method (SVM without any DA). More importantly, some well-performing DA methods, e.g., GFK~\cite{gong2012geodesic}, DRFS-T~\cite{zong2018domain}, and the proposed RSTR, have significant improvements of at least 0.1000 (average $mean$ $F1$-$score$) and 10\% (average $Accuracy$) compared with the baseline results in either TYPE-I or TYPE-II experiments. Based on the above observations, we are able to reach the conclusion that considering CDMER as an DA problem is no doubt an effective solution for the CDMER problem. It is a good choice to develop excellent DA methods to relieve the feature distribution mismatch between the samples from different micro-expression databases.

Moreover, we observe that our proposed RSTR achieves the best average results among all the methods in the CDMER experiments. Specifically, RSTR achieves the average $mean$ $F1$-$score$ / $Accuracy$ of 0.7381 / 73.98\% in TYPE-I experiments and 0.5587 / 57.74\% for TYPE-II experiments, which are significantly higher than most of DA methods. The superior performance of RSTR may attribute to the consideration of the facial region information in its design. Let us recall the spatial division scheme in Fig.~\ref{fig:fig2} adopted by our benchmark, which is used together with spatiotemporal descriptors for describing micro-expressions. In fact, the facial local regions yielded by such spatial division scheme have different contributions to distinguish micro-expressions. Such different contributions are often neglected in most works of micro-expression analysis. Recent work in~\cite{zong2018learning} had shown that it is beneficial for micro-expression recognition if we consider the different contributions of different facial regions. From the advantages of RSTR, it can convincingly verify that taking into consideration the positive cues associated with distinguishing micro-expressions, e.g., different contributions of different facial regions, offers us a new insight to develop the DA methods such that these DA methods are more applicable to CDMER problem.

\subsubsection{Results of CDMER Experiments by Using Micro-Expression Features}

The detailed $mean$ $F1$-$score$ / $Accuracy$ achieved by various micro-expression features are depicted in Tables~\ref{tab:tab5} and~\ref{tab:tab6}. From these two tables, it is clear to see that LPQ-TOP with $decorr = 0$ and LBP-SIP with $R=3$ achieve the best average performance in terms of $mean$ $F1$-$score$ and $Accuracy$ in TYPE-I and TYPE-II CDMER experiments, respectively. More importantly, we notice that LPQ-TOP$_{(decorr = 0)}$ obtains the average $mean$ $F1$-$score$ / $Accuracy$ of 0.6325 / 64.05\% in the first type of experiments, which are even competitive among the results of DA methods. By referring Table~\ref{tab:tab3} and Tables~\ref{tab:tab5}, we can find that LPQ-TOP$_{(decorr = 0)}$ outperforms TCA (0.6238 / 64.01\%) and is very competitive against STM (0.6440 / 65.78\%) and DRLS (0.6552 / 66.60\%). In addition, we also observe that several micro-expression features achieve very promising results. For example, LPQ-TOP$_{(decorr=0.1)}$ achieves the $mean$ $F1$-$score$ of 0.9455 and $Accuracy$ of 94.37\% in Exp.1, which are significantly better than all the DA methods. In Exp.9 and Exp.10, LBP-SIP$_{(R=3)}$ is a very good competitor against DA methods. Based on the above observations, it is believed that developing database-invariant spatiotemporal descriptors for robustly describing micro-expressions also provide a promising and feasible way to solve the CDMER problem.

However, we have to admit that most of the existing spatiotemporal descriptors including the above well-performing ones are still not satisfactory and cannot completely meet the requirement in CDMER problem. More specifically, several limitations still exist in current micro-expression feature research. Firstly, the performance of most spatiotemporal descriptors is not stable. In other words, one spatiotemporal descriptor performs well in Task A but very poor in Task B. An example is LPQ-TOP. It can be seen that LPQ-TOP with two different parameters both achieve satisfactory results beating the baseline method. But the performance of both two LBP-TOP based micro-expression features decrease sharply and even under the level of baseline method. Secondly, it is clear to see that the performance of most of handcrafted spatiotemporal descriptors is strongly affected by its parameters. As the results of LBP-TOP showed, its performance varies very sensitively in nearly all the experiments of our CDMER benchmark with the changes of its parameter $R$ and $P$. Therefore, in the future study of spatiotemporal descriptors used for describing micro-expressions, we should also consider to reduce the sensitiveness of its parameters such that it will be simpler and more convenient to use in dealing with the CDMER problem. Finally, it should be pointed out that at present it is not enough for solving CDMER problem to simply use existing spatiotemporal descriptors as micro-expression features because its average performance is still far from the DA methods. From another point of view, it is believed that there is still very large development space in this direction, i.e., developing robust (database-invariant) micro-expression features.

\begin{table}[!t]
\scriptsize
\renewcommand{\arraystretch}{1.2}
\caption{Experimental results (mean F1-score / Accuracy) of finetuned C3D features.}
\label{tab:tab7}
\centering
\begin{tabular}{|l|c|c|c|}
\hline
Deep Features & Exp.1: H $\rightarrow$ V & Exp.3: H $\rightarrow$ N & Exp.8: H $\rightarrow$ C \\\hline\hline
Baseline & 0.8002 / 80.28 & 0.5455 / 53.52 & 0.3245 / 48.46 \\\hline
TCA & 0.8269 / 83.10 & 0.5828 / 59.15 & 0.4870 / 53.08 \\\hline
C3D-FC1 (Sports1M) & 0.1577 / 30.99 & 0.1667 / 30.99 & 0.2394 / 56.15 \\\hline
C3D-FC2 (Sports1M) & 0.2555 / 36.62 & 0.2804 / 33.80 & 0.1317 / 24.62\\\hline
C3D-FC1 (Finetune) & 0.4858 / 56.34 & 0.2751 / 35.21 & 0.3276 / 34.62 \\\hline
C3D-FC2 (Finetune) & 0.4144 / 47.89 & 0.3758 / 40.85 & 0.3390 / 36.92 \\\hline
\end{tabular}
\end{table}

Lastly, we discuss the deep features evaluated in the benchmark. From the results of deep features extracted by C3D pretrained on Sports1M and UCF101, we can observe that these four deep features perform poorly in nearly all the CDMER experiments and cannot reach the level of most handcrafted features. In fact, this is foreseeable because the Sports1M and UCF101 served for C3D pretraining are both action databases, whose samples are quite different from the micro-expression ones. In fact, to enable the C3D to gain the micro-expression information, we can finetune the pretrained models based on the source database and then use the finetuned model to serve as feature extractor. To this end, we finetune the C3D pretrained on Sports1M based on SMIC (HS) and then extract the deep features to conduct the experiments of using SMIC (HS) as source database including Exp.1, Exp.3, and Exp.8. Note that for finetuning, we augment the samples in SMIC (HS) to 3350\footnote{Specifically, we first use TIM~\cite{zhou2011towards} to normalize the frame number of all the micro-expression samples to 160. Then we divide these 160 frames into 10 parts without overlap. Finally, we randomly extract one frame from all the parts to obtain a micro-expression sample with frame length of 16.} and divide the samples into a training set whose sample number is 2550 and a validation set containing 800 samples. The experimental results are shown in Table~\ref{tab:tab7}, where we also list the results of C3D pretrained on Sports1M, the baseline method (LBP-TOP with $R=3,P=8$), and TCA (a representative DA method). From Table~\ref{tab:tab7}, it is clear that by resorting to finetuning strategy, the performance of C3D features can be improved significantly compared with the original model. It is also interesting to see that in Exp.8, the deep features extracted by finetuned C3D even outperform the baseline method (LBP-TOP with $R=3,P=8$) in term of mean F1-score. However, compared with TCA (a representative DA method) and the baseline method, the performance of deep features is still very poor and not satisfactory. We think there may be two possible reasons. Firstly, it may attribute to the problem of lacking enough good data for finetuning the C3D models. It is clear that the training or finetuning of deep learning models requires large numbers of samples while the sample numbers of the CASME~II and SMIC are too small. Although we can use some methods to augment the samples, the augmented micro-expression samples are not satisfactory. Therefore, more micro-expression samples need be collected such that a better-performing C3D can be finetuned to deal with CDMER problem. The second reason may be that it seems not enough for learning the database-invariant deep features to simply finetune the deep model based on the source database. Leveraging the idea of domain adaptation methods, i.e., reducing the difference between source and target domains, to finetune deep models may offer a feasible solution to improve the performance of deep features.

\section{Conclusion}
\label{conclusion}
In this paper, we have investigated the cross-database micro-expression recognition (CDMER) problem by conducting a standard benchmark evaluation from two different perspectives including domain adaptation (DA) and spatiotemporal features used for describing micro-expressions. First, under a well-designed evaluation protocol, we make use of two widely-used micro-expression databases, i.e., CASME~II and SMIC, to set up two types of CDMER experiments. Then, we collect NINE state-of-the-art DA methods and SIX excellent spatiotemporal descriptors to perform benchmark evaluation based on our designed CDMER protocol. Moreover, we also propose a novel DA method called region selective transfer regression (RSTR), which fully considers the different contributions of the facial local regions in recognizing micro-expressions. Finally, comprehensive discussions for the benchmark evaluation results are provided. More importantly, we reach a useful conclusion from the comparison between the proposed RSTR and other state-of-the-art DA methods that it is beneficial for dealing with the CDMER problem if we take full advantage of important cues associated with micro-expressions like facial local region information considered in the design of our RSTR. In addition, all the collected data and codes involving the benchmark evaluation in our paper will be released as soon as possible. We hope our work can advance the micro-expression analysis research by inspiring more researchers to focus on the CDMER problem. In the future, we will insist to collect more well-performing DA methods and micro-expression features for updating the evaluation results in our project page.

\ifCLASSOPTIONcaptionsoff
\newpage
\fi



\bibliographystyle{IEEEtran}
\bibliography{IEEETrans2019-20191111}

\begin{thebibliography}{10}
\providecommand{\url}[1]{#1}
\csname url@samestyle\endcsname
\providecommand{\newblock}{\relax}
\providecommand{\bibinfo}[2]{#2}
\providecommand{\BIBentrySTDinterwordspacing}{\spaceskip=0pt\relax}
\providecommand{\BIBentryALTinterwordstretchfactor}{4}
\providecommand{\BIBentryALTinterwordspacing}{\spaceskip=\fontdimen2\font plus
\BIBentryALTinterwordstretchfactor\fontdimen3\font minus
  \fontdimen4\font\relax}
\providecommand{\BIBforeignlanguage}[2]{{%
\expandafter\ifx\csname l@#1\endcsname\relax
\typeout{** WARNING: IEEEtran.bst: No hyphenation pattern has been}%
\typeout{** loaded for the language `#1'. Using the pattern for}%
\typeout{** the default language instead.}%
\else
\language=\csname l@#1\endcsname
\fi
#2}}
\providecommand{\BIBdecl}{\relax}
\BIBdecl

\bibitem{ekman1969nonverbal}
P.~Ekman and W.~V. Friesen, ``Nonverbal leakage and clues to deception,''
  \emph{Psychiatry}, vol.~32, no.~1, pp. 88--106, 1969.

\bibitem{haggard1966micromomentary}
E.~A. Haggard and K.~S. Isaacs, ``Micromomentary facial expressions as
  indicators of ego mechanisms in psychotherapy,'' in \emph{Methods of research
  in psychotherapy}.\hskip 1em plus 0.5em minus 0.4em\relax Springer, 1966, pp.
  154--165.

\bibitem{ekman2009telling}
P.~Ekman, \emph{Telling lies: Clues to deceit in the marketplace, politics, and
  marriage (revised edition)}.\hskip 1em plus 0.5em minus 0.4em\relax WW Norton
  \&amp; Company, 2009.

\bibitem{pfister2011recognising}
T.~Pfister, X.~Li, G.~Zhao, and M.~Pietik{\"a}inen, ``Recognising spontaneous
  facial micro-expressions,'' in \emph{International Conference on Computer
  Vision}.\hskip 1em plus 0.5em minus 0.4em\relax IEEE, 2011, pp. 1449--1456.

\bibitem{wang2014lbp}
Y.~Wang, J.~See, R.~C.-W. Phan, and Y.-H. Oh, ``Lbp with six intersection
  points: Reducing redundant information in lbp-top for micro-expression
  recognition,'' in \emph{Asian Conference on Computer Vision}.\hskip 1em plus
  0.5em minus 0.4em\relax Springer, 2014, pp. 525--537.

\bibitem{wang2015micro}
S.-J. Wang, W.-J. Yan, X.~Li, G.~Zhao, C.-G. Zhou, X.~Fu, M.~Yang, and J.~Tao,
  ``Micro-expression recognition using color spaces,'' \emph{IEEE Transactions
  on Image Processing}, vol.~24, no.~12, pp. 6034--6047, 2015.

\bibitem{liu2016main}
Y.-J. Liu, J.-K. Zhang, W.-J. Yan, S.-J. Wang, G.~Zhao, and X.~Fu, ``A main
  directional mean optical flow feature for spontaneous micro-expression
  recognition,'' \emph{IEEE Transactions on Affective Computing}, vol.~7,
  no.~4, pp. 299--310, 2016.

\bibitem{kim2016micro}
D.~H. Kim, W.~J. Baddar, and Y.~M. Ro, ``Micro-expression recognition with
  expression-state constrained spatio-temporal feature representations,'' in
  \emph{Proceedings of the 2016 ACM on Multimedia Conference}.\hskip 1em plus
  0.5em minus 0.4em\relax ACM, 2016, pp. 382--386.

\bibitem{lu2016micro}
P.~Lu, W.~Zheng, Z.~Wang, Q.~Li, Y.~Zong, M.~Xin, and L.~Wu, ``Micro-expression
  recognition by regression model and group sparse spatio-temporal feature
  learning,'' \emph{IEICE TRANSACTIONS on Information and Systems}, vol.~99,
  no.~6, pp. 1694--1697, 2016.

\bibitem{xu2017microexpression}
F.~Xu, J.~Zhang, and J.~Z. Wang, ``Microexpression identification and
  categorization using a facial dynamics map,'' \emph{IEEE Transactions on
  Affective Computing}, vol.~8, no.~2, pp. 254--267, 2017.

\bibitem{happy2017fuzzy}
S.~Happy and A.~Routray, ``Fuzzy histogram of optical flow orientations for
  micro-expression recognition,'' \emph{IEEE Transactions on Affective
  Computing}, 2017.

\bibitem{zong2018learning}
Y.~Zong, X.~Huang, W.~Zheng, Z.~Cui, and G.~Zhao, ``Learning from hierarchical
  spatiotemporal descriptors for micro-expression recognition,'' \emph{IEEE
  Transactions on Multimedia}, 2018.

\bibitem{zhao2007dynamic}
G.~Zhao and M.~Pietik{\"a}inen, ``Dynamic texture recognition using local
  binary patterns with an application to facial expressions,'' \emph{IEEE
  Transactions on Pattern Analysis and Machine Intelligence}, vol.~29, no.~6,
  pp. 915--928, 2007.

\bibitem{ekman1997face}
P.~Ekman and E.~L. Rosenberg, \emph{What the face reveals: Basic and applied
  studies of spontaneous expression using the Facial Action Coding System
  (FACS)}.\hskip 1em plus 0.5em minus 0.4em\relax Oxford University Press, USA,
  1997.

\bibitem{krizhevsky2012imagenet}
A.~Krizhevsky, I.~Sutskever, and G.~E. Hinton, ``Imagenet classification with
  deep convolutional neural networks,'' in \emph{Advances in Neural Information
  Processing Systems}, 2012, pp. 1097--1105.

\bibitem{hochreiter1997long}
S.~Hochreiter and J.~Schmidhuber, ``Long short-term memory,'' \emph{Neural
  computation}, vol.~9, no.~8, pp. 1735--1780, 1997.

\bibitem{schuller2010cross}
B.~Schuller, B.~Vlasenko, F.~Eyben, M.~Wollmer, A.~Stuhlsatz, A.~Wendemuth, and
  G.~Rigoll, ``Cross-corpus acoustic emotion recognition: Variances and
  strategies,'' \emph{IEEE Transactions on Affective Computing}, vol.~1, no.~2,
  pp. 119--131, 2010.

\bibitem{yan2016cross}
K.~Yan, W.~Zheng, Z.~Cui, and Y.~Zong, ``Cross-database facial expression
  recognition via unsupervised domain adaptive dictionary learning,'' in
  \emph{International Conference on Neural Information Processing}.\hskip 1em
  plus 0.5em minus 0.4em\relax Springer, 2016, pp. 427--434.

\bibitem{zheng2016personalizing}
W.-L. Zheng and B.-L. Lu, ``Personalizing eeg-based affective models with
  transfer learning,'' in \emph{Proceedings of the Twenty-Fifth International
  Joint Conference on Artificial Intelligence}.\hskip 1em plus 0.5em minus
  0.4em\relax AAAI Press, 2016, pp. 2732--2738.

\bibitem{pan2010survey}
S.~J. Pan and Q.~Yang, ``A survey on transfer learning,'' \emph{IEEE
  Transactions on Knowledge and Data Engineering}, vol.~22, no.~10, pp.
  1345--1359, 2010.

\bibitem{zong2017learning}
Y.~Zong, X.~Huang, W.~Zheng, Z.~Cui, and G.~Zhao, ``Learning a target sample
  re-generator for cross-database micro-expression recognition,'' in
  \emph{Proceedings of the 2017 ACM on Multimedia Conference}.\hskip 1em plus
  0.5em minus 0.4em\relax ACM, 2017, pp. 872--880.

\bibitem{zong2018domain}
Y.~Zong, W.~Zheng, X.~Huang, J.~Shi, Z.~Cui, and G.~Zhao, ``Domain regeneration
  for cross-database micro-expression recognition,'' \emph{IEEE Transactions on
  Image Processing}, vol.~27, no.~5, pp. 2484--2498, 2018.

\bibitem{zong2019cross}
Y.~Zong, W.~Zheng, X.~Hong, C.~Tang, Z.~Cui, and G.~Zhao, ``Cross-database
  micro-expression recognition: A benchmark,'' in \emph{Proceedings of the 2019
  on International Conference on Multimedia Retrieval}.\hskip 1em plus 0.5em
  minus 0.4em\relax ACM, 2019, pp. 354--363.

\bibitem{yan2014casme}
W.-J. Yan, X.~Li, S.-J. Wang, G.~Zhao, Y.-J. Liu, Y.-H. Chen, and X.~Fu,
  ``Casme {II}: An improved spontaneous micro-expression database and the
  baseline evaluation,'' \emph{PloS one}, vol.~9, no.~1, p. e86041, 2014.

\bibitem{li2013spontaneous}
X.~Li, T.~Pfister, X.~Huang, G.~Zhao, and M.~Pietik{\"a}inen, ``A spontaneous
  micro-expression database: Inducement, collection and baseline,'' in
  \emph{Automatic face and gesture recognition (fg), 2013 10th ieee
  international conference and workshops on}.\hskip 1em plus 0.5em minus
  0.4em\relax IEEE, 2013, pp. 1--6.

\bibitem{le2014spontaneous}
A.~C. Le~Ngo, R.~C.-W. Phan, and J.~See, ``Spontaneous subtle expression
  recognition: Imbalanced databases and solutions,'' in \emph{Proceedings of
  the 12th Asian Conference on Computer Vision (ACCV)}.\hskip 1em plus 0.5em
  minus 0.4em\relax Springer, 2014, pp. 33--48.

\bibitem{oh2015monogenic}
Y.-H. Oh, A.~C. Le~Ngo, J.~See, S.-T. Liong, R.~C.-W. Phan, and H.-C. Ling,
  ``Monogenic riesz wavelet representation for micro-expression recognition,''
  in \emph{Proceedings of the 20th IEEE International Conference on Digital
  Signal Processing (DSP)}.\hskip 1em plus 0.5em minus 0.4em\relax IEEE, 2015,
  pp. 1237--1241.

\bibitem{le2016sparsity}
A.~C. Le~Ngo, J.~See, and C.-W.~R. Phan, ``Sparsity in dynamics of spontaneous
  subtle emotion: Analysis \& application,'' \emph{IEEE Transactions on
  Affective Computing}, 2016.

\bibitem{le2016eulerian}
A.~C. Le~Ngo, Y.-H. Oh, R.~C.-W. Phan, and J.~See, ``Eulerian emotion
  magnification for subtle expression recognition,'' in \emph{Proceedings of
  the 41st IEEE International Conference on Acoustics, Speech and Signal
  Processing (ICASSP)}.\hskip 1em plus 0.5em minus 0.4em\relax IEEE, 2016, pp.
  1243--1247.

\bibitem{zhou2011towards}
Z.~Zhou, G.~Zhao, and M.~Pietik{\"a}inen, ``Towards a practical lipreading
  system,'' in \emph{Computer Vision and Pattern Recognition (CVPR), 2011 IEEE
  Conference on}.\hskip 1em plus 0.5em minus 0.4em\relax IEEE, 2011, pp.
  137--144.

\bibitem{chang2011libsvm}
C.-C. Chang and C.-J. Lin, ``Libsvm: a library for support vector machines,''
  \emph{ACM transactions on intelligent systems and technology (TIST)}, vol.~2,
  no.~3, p.~27, 2011.

\bibitem{hassan2013acoustic}
A.~Hassan, R.~Damper, and M.~Niranjan, ``On acoustic emotion recognition:
  compensating for covariate shift,'' \emph{IEEE Transactions on Audio, Speech,
  and Language Processing}, vol.~21, no.~7, pp. 1458--1468, 2013.

\bibitem{huang2006correcting}
J.~Huang, A.~Gretton, K.~M. Borgwardt, B.~Sch{\"o}lkopf, and A.~J. Smola,
  ``Correcting sample selection bias by unlabeled data,'' in \emph{Advances in
  Neural Information Processing Systems}, 2006, pp. 601--608.

\bibitem{kanamori2009least}
T.~Kanamori, S.~Hido, and M.~Sugiyama, ``A least-squares approach to direct
  importance estimation,'' \emph{The Journal of Machine Learning Research},
  vol.~10, pp. 1391--1445, 2009.

\bibitem{sugiyama2008direct}
M.~Sugiyama, S.~Nakajima, H.~Kashima, P.~V. Buenau, and M.~Kawanabe, ``Direct
  importance estimation with model selection and its application to covariate
  shift adaptation,'' in \emph{Advances in Neural Information Processing
  Systems}, 2008, pp. 1433--1440.

\bibitem{pan2011domain}
S.~J. Pan, I.~W. Tsang, J.~T. Kwok, and Q.~Yang, ``Domain adaptation via
  transfer component analysis,'' \emph{IEEE Transactions on Neural Networks},
  vol.~22, no.~2, pp. 199--210, 2011.

\bibitem{gong2012geodesic}
B.~Gong, Y.~Shi, F.~Sha, and K.~Grauman, ``Geodesic flow kernel for
  unsupervised domain adaptation,'' in \emph{Computer Vision and Pattern
  Recognition (CVPR), 2012 IEEE Conference on}.\hskip 1em plus 0.5em minus
  0.4em\relax IEEE, 2012, pp. 2066--2073.

\bibitem{fernando2013unsupervised}
B.~Fernando, A.~Habrard, M.~Sebban, and T.~Tuytelaars, ``Unsupervised visual
  domain adaptation using subspace alignment,'' in \emph{Proceedings of the
  IEEE international conference on computer vision}, 2013, pp. 2960--2967.

\bibitem{chu2013selective}
W.-S. Chu, F.~De~la Torre, and J.~F. Cohn, ``Selective transfer machine for
  personalized facial action unit detection,'' in \emph{Proceedings of the IEEE
  Conference on Computer Vision and Pattern Recognition}, 2013, pp. 3515--3522.

\bibitem{chu2017selective}
------, ``Selective transfer machine for personalized facial expression
  analysis,'' \emph{IEEE transactions on pattern analysis and machine
  intelligence}, vol.~39, no.~3, pp. 529--545, 2017.

\bibitem{long2015domain}
M.~Long, J.~Wang, J.~Sun, and S.~Y. Philip, ``Domain invariant transfer kernel
  learning,'' \emph{IEEE Transactions on Knowledge and Data Engineering},
  vol.~27, no.~6, pp. 1519--1532, 2015.

\bibitem{paivarinta2011volume}
J.~P{\"a}iv{\"a}rinta, E.~Rahtu, and J.~Heikkil{\"a}, ``Volume local phase
  quantization for blur-insensitive dynamic texture classification,'' in
  \emph{Scandinavian Conference on Image Analysis}.\hskip 1em plus 0.5em minus
  0.4em\relax Springer, 2011, pp. 360--369.

\bibitem{ojansivu2008blur}
V.~Ojansivu and J.~Heikkil{\"a}, ``Blur insensitive texture classification
  using local phase quantization,'' in \emph{International conference on image
  and signal processing}.\hskip 1em plus 0.5em minus 0.4em\relax Springer,
  2008, pp. 236--243.

\bibitem{li2017towards}
X.~Li, X.~Hong, A.~Moilanen, X.~Huang, T.~Pfister, G.~Zhao, and
  M.~Pietik{\"a}inen, ``Towards reading hidden emotions: A comparative study of
  spontaneous micro-expression spotting and recognition methods,'' \emph{IEEE
  Transactions on Affective Computing}, 2017.

\bibitem{dalal2005histograms}
N.~Dalal and B.~Triggs, ``Histograms of oriented gradients for human
  detection,'' in \emph{Computer Vision and Pattern Recognition, 2005. CVPR
  2005. IEEE Computer Society Conference on}, vol.~1.\hskip 1em plus 0.5em
  minus 0.4em\relax IEEE, 2005, pp. 886--893.

\bibitem{tran2015learning}
D.~Tran, L.~Bourdev, R.~Fergus, L.~Torresani, and M.~Paluri, ``Learning
  spatiotemporal features with 3d convolutional networks,'' in \emph{Computer
  Vision (ICCV), 2015 IEEE International Conference on}.\hskip 1em plus 0.5em
  minus 0.4em\relax IEEE, 2015, pp. 4489--4497.

\bibitem{simonyan2014very}
K.~Simonyan and A.~Zisserman, ``Very deep convolutional networks for
  large-scale image recognition,'' \emph{arXiv preprint arXiv:1409.1556}, 2014.

\bibitem{borgwardt2006integrating}
K.~M. Borgwardt, A.~Gretton, M.~J. Rasch, H.-P. Kriegel, B.~Sch{\"o}lkopf, and
  A.~J. Smola, ``Integrating structured biological data by kernel maximum mean
  discrepancy,'' \emph{Bioinformatics}, vol.~22, no.~14, pp. e49--e57, 2006.

\bibitem{scholkopf1998nonlinear}
B.~Sch{\"o}lkopf, A.~Smola, and K.-R. M{\"u}ller, ``Nonlinear component
  analysis as a kernel eigenvalue problem,'' \emph{Neural computation},
  vol.~10, no.~5, pp. 1299--1319, 1998.

\bibitem{zheng2014multi}
W.~Zheng, ``Multi-view facial expression recognition based on group sparse
  reduced-rank regression,'' \emph{IEEE Transactions on Affective Computing},
  vol.~5, no.~1, pp. 71--85, 2014.

\bibitem{lin2010augmented}
Z.~Lin, M.~Chen, and Y.~Ma, ``The augmented lagrange multiplier method for
  exact recovery of corrupted low-rank matrices,'' \emph{arXiv preprint
  arXiv:1009.5055}, 2010.

\bibitem{liu2009slep}
J.~Liu, S.~Ji, and J.~Ye, ``Slep: Sparse learning with efficient projections,''
  \emph{Arizona State University}, vol.~6, p. 491, 2009.

\bibitem{ding2014latent}
Z.~Ding, M.~Shao, and Y.~Fu, ``Latent low-rank transfer subspace learning for
  missing modality recognition.'' in \emph{AAAI}, 2014, pp. 1192--1198.

\bibitem{al2014supervised}
M.~Al-Shedivat, J.~J.-Y. Wang, M.~Alzahrani, J.~Z. Huang, and X.~Gao,
  ``Supervised transfer sparse coding,'' in \emph{Proceedings of the
  Twenty-Eighth AAAI conference on artificial intelligence}.\hskip 1em plus
  0.5em minus 0.4em\relax The AAAI Press, 2014.

\bibitem{long2014transfer}
M.~Long, J.~Wang, G.~Ding, J.~Sun, and P.~S. Yu, ``Transfer joint matching for
  unsupervised domain adaptation,'' in \emph{Proceedings of the IEEE conference
  on computer vision and pattern recognition}, 2014, pp. 1410--1417.

\bibitem{wold1987principal}
S.~Wold, K.~Esbensen, and P.~Geladi, ``Principal component analysis,''
  \emph{Chemometrics and intelligent laboratory systems}, vol.~2, no. 1-3, pp.
  37--52, 1987.

\bibitem{hong2016lbp}
X.~Hong, Y.~Xu, and G.~Zhao, ``Lbp-top: a tensor unfolding revisit,'' in
  \emph{Asian Conference on Computer Vision}.\hskip 1em plus 0.5em minus
  0.4em\relax Springer, 2016, pp. 513--527.

\bibitem{karpathy2014large}
A.~Karpathy, G.~Toderici, S.~Shetty, T.~Leung, R.~Sukthankar, and L.~Fei-Fei,
  ``Large-scale video classification with convolutional neural networks,'' in
  \emph{Proceedings of the IEEE conference on Computer Vision and Pattern
  Recognition}, 2014, pp. 1725--1732.

\bibitem{soomro2012ucf101}
K.~Soomro, A.~R. Zamir, and M.~Shah, ``Ucf101: A dataset of 101 human actions
  classes from videos in the wild,'' \emph{arXiv preprint arXiv:1212.0402},
  2012.

\end{thebibliography}
\end{document}